\documentclass{article}

\usepackage{arxiv}

\usepackage[utf8]{inputenc} 
\usepackage[T1]{fontenc}    
\usepackage{hyperref}       
\usepackage{url}            
\usepackage{booktabs}       
\usepackage{amsfonts}       
\usepackage{amssymb}
\usepackage{amsmath}
\usepackage{nicefrac}       
\usepackage{microtype}      
\usepackage{lipsum}		
\usepackage{graphicx}
\usepackage{natbib}
\usepackage{doi}
\usepackage{multirow}
\usepackage{siunitx}
\DeclareSIUnit{\celsius}{\SIUnitSymbolDegree C}
\DeclareSIUnit{\kjmm}{\kilo\joule\per\meter\squared}
\usepackage{algorithm} 
\usepackage{algpseudocode} 
\usepackage{authblk}

\algdef{SE}
[CLASS]
{Class}
{EndClass}
[1]
{\textbf{class} \textsc{#1}}
{\textbf{end class}}

\title{Reinforcement Learning Constrained Beam Search for Parameter Optimization of Paper Drying under Flexible Constraints}

\author[1]{\href{https://orcid.org/0000-0003-1150-5138}{Siyuan Chen}}
\author[2]{\href{https://orcid.org/0000-0002-5381-5399}{Hanshen Yu}}
\author[2]{\href{https://orcid.org/0000-0002-6378-475X}{Jamal Yagoobi}}
\author[1, 3]{\href{https://orcid.org/0000-0002-3299-2222}{Chenhui Shao}}

\affil[1]{\footnotesize Department of Mechanical Science and Engineering, University of Illinois at Urbana-Champaign, Urbana, IL 61801}
\affil[2]{\footnotesize Department of Mechanical and Materials Engineering, Worcester Polytechnic Institute, Worcester, MA 01609}
\affil[3]{\footnotesize Department of Mechanical Engineering, University of Michigan, Ann Arbor, MI 48109}

\affil[ ]{%
  \footnotesize
  \texttt{siyuanc2@illinois.edu, hyu5@wpi.edu, jyagoobi@wpi.edu, chshao@umich.edu}
}

\renewcommand{\headeright}{Working Paper}
\renewcommand{\undertitle}{Working Paper}
\renewcommand{\shorttitle}{RLCBS for Paper Drying}

\hypersetup{
pdftitle={Reinforcement Learning Constrained Beam Search for Parameter Optimization of Paper Drying under Flexible Constraints},
pdfsubject={cs.LG, cs.SY},
pdfauthor={Siyuan Chen, Hanshen Yu, Jamal Yagoobi, Chenhui Shao},
pdfkeywords={Reinforcement learning, Constrained beam search, Drying, Papermaking, Process optimization},
}

\begin{document}

\maketitle

\begin{abstract}
Existing approaches to enforcing design constraints in Reinforcement Learning (RL) applications often rely on training-time penalties in the reward function or training/inference-time invalid action masking, but these methods either cannot be modified after training, or are limited in the types of constraints that can be implemented. To address this limitation, we propose Reinforcement Learning Constrained Beam Search (RLCBS) for inference-time refinement in combinatorial optimization problems. This method respects flexible, inference-time constraints that support exclusion of invalid actions and forced inclusion of desired actions, and employs beam search to maximize sequence probability for more sensible constraint incorporation. RLCBS is extensible to RL-based planning and optimization problems that do not require real-time solution, and we apply the method to optimize process parameters for a novel modular testbed for paper drying. An RL agent is trained to minimize energy consumption across varying machine speed levels by generating optimal dryer module and air supply temperature configurations. Our results demonstrate that RLCBS outperforms NSGA-II under complex design constraints on drying module configurations at inference-time, while providing a 2.58-fold or higher speed improvement. \let\thefootnote\relax\footnotetext{Corresponding author: Chenhui Shao.}
\end{abstract}

\keywords{Reinforcement~learning \and Constrained~beam~search \and Drying \and Papermaking \and Process~optimization}

\section{Introduction}
Many games and optimization problems feature complex rules or design constraints that limit how Reinforcement Learning (RL) agents interact with the environment. In recent literature, there exist two common approaches to ensure compliance with these constraints: (1) Imposing penalties in the reward function when the RL agent violates a design constraint during training~\citep{yang2023automatic}, and (2) ``Masking out'' invalid actions at training and/or inference-time~\citep{vinyals2019grandmaster, ye2020mastering} by setting the log-probabilities of invalid actions to $-\infty$ for a policy-based RL agent operating on discrete action spaces. However, both approaches have limitations in real-world scenarios.

The first approach, integrating constraints in the reward function, offers good flexibility in constraint types, as it is possible to penalize undesired actions and promote desired ones through the reward function. However, this implies that all constraints are to be encoded in the reward function before training, and changes in design constraints post-training can lead to suboptimal solutions, with the RL agent either not exploring as close to the constraint boundary as possible or consistently violating more stringent constraints. Consequently, changes in design constraints necessitate fine-tuning or retraining the model with an updated reward function. This behavior hinders real-world applications, which will be demonstrated in our case study of optimizing the process configuration and control parameters of a modular dryer testbed. Here, constraints may change frequently based on the number of drying modules available, which requires rapid adaptation without the time allotted for additional training. Moreover, this method is found to scale poorly as the number of invalid actions increases, making it difficult for the RL algorithm to achieve positive rewards and make progress during training when a large proportion of the action space is invalid~\citep{huang2020closer}.

The second approach, invalid action masking, allows for changes to the action masking logic even at inference-time, making it possible to rapidly adapt to design constraint changes. However, this method is limited in the types of constraint that can be applied. For example, it may be difficult to guarantee that a specific number of desired actions appear in the RL-generated action sequence, especially when the desired actions are not preferred by the trained RL policy. Although one could mask out all other actions or manipulate the RL policy-generated action probability distribution to promote desired actions, it cannot be guaranteed that the constraints are incorporated in the generated actions in the most sensible way, i.e., resulting in the least impact to the cumulative episodic reward.

In consideration of the limitations of the two existing methods for RL constraint compliance discussed above, there is a need for optimized placement of more flexible design constraints during RL inference time. Chen et al. showed that it is possible to apply beam search, a popular approach for natural language generation, to RL action generation, provided that the simulation environment is deterministic and that real-time performance is not necessary. Following this trail, we explored the use of constrained beam search (CBS) for policy-based RL operating on discrete action spaces. First proposed by~\citep{anderson-etal-2017-guided} to force the inclusion of recognized objects in model-generated image captions, CBS has been proposed as a method to incorporate additional knowledge into model output without requiring modification of model parameters or training data~\citep{hokamp-liu-2017-lexically}. CBS have been widely adopted in various natural language processing libraries~\citep{ott-etal-2019-fairseq, wolf-etal-2020-transformers} to force the inclusion or exclusion of specific words or phrases in text generated by large language models. This approach offers a promising solution to the challenge of flexible constraint implementation in RL, as it not only provides logits processors that can be used for invalid action masking similar to Maskable~PPO~\citep{tang2020implementing}, but also introduces \emph{lexical} and \emph{phrasal} constraints that serve as foundations to actively promote the inclusion of desired actions in RL-generated action sequences. To ensure optimal placement of desired actions in the generated sequence, CBS allows for maintaining a large number of beams spanning different stages of constraint fulfillment. Consequently, the constraints are fulfilled in the most \emph{sensible} way by returning the beam with the highest cumulative action log-probabilities (beam scores) at the end of the generation process.

In this paper, we describe a constrained beam search (CBS) algorithm for policy-based RL models operating on discrete action spaces. Named Reinforcement Learning-Guided Constrained Beam Search (RLCBS), the algorithm is based on a CBS implementation from the Huggingface Transformers library~\citep{wolf-etal-2020-transformers} for natural language processing and can be easily incorporated to policy-based RL algorithms implemented in the Stable-baselines3~\citep{stable-baselines3} library. We apply RLCBS to the task of optimizing process parameters of a modular, reconfigurable, fully electric test bed for drying of pulp and paper. A Proximal Policy Optimization (PPO)~\citep{schulman2017proximal} RL agent is trained to generate optimal dryer module configurations and air temperature for paper drying with minimized overall energy consumption, while adapting to varying machine speed levels. We show that the RLCBS algorithm allows the RL agent to match or outperform NSGA-II under multiple design constraints on the configuration of dryer modules unseen during RL training, while providing an average 2.58-fold speed advantage.

The research gaps this work seeks to address are as follows.
\begin{itemize}
    \item To our knowledge, no previous work discussed the application of beam search in RL that supports \emph{positive} constraints, such that desired actions can be included in the generated sequence in the most sensible way even if they are not preferred by the RL policy. Existing literature focused on \emph{negative} constraints that prevents certain actions from being generated.
    \item The case study presented in this work features a novel, modular, and reconfigurable testbed for paper drying.
    \item Few existing literature has discussed optimization of drying processes under continuously variable process operating conditions while responding to changes in design constraints.
\end{itemize}

The remainder of this paper is organized as follows. We describe the RLCBS algorithm and the background of previous related work in Section~\ref{sec:algorithm}. Section~\ref{sec:case_study} presents the physical dryer testbed used for the case study, the theoretical model, the RL problem setup, and practical implementation notes of RLCBS on the case study. We present a comparison of RLCBS and NSGA-II under constraints in Section~\ref{sec:results}, and a discussion on the results is provided in Section~\ref{sec:discussion}. Finally, Section~\ref{sec:conclusion} concludes the paper.

We plan to release the RLCBS code as an extension to RLGBS at \url{https://github.com/siyuanc2/RLGBS}. Although the source code for drying simulation cannot be publicly released, we plan to release a docker container with a compiled binary of the drying simulation environment for experiment reproducibility.

\section{Method}
\label{sec:algorithm}
\subsection{Background}
\subsubsection{Beam Search for RL in Combinatorial Optimization} For a stochastic RL policy $\pi$ parameterized by weights $\theta$ and with a discrete action space $\mathcal{A}$, given the current state of the environment $s_t$ at timestep $t$, the policy network outputs a categorical distribution across the action space:
\begin{equation}
\pi_\theta(s_t) = \{p_\theta(a|s_t)  \forall a \in \mathcal{A}\}.
\end{equation}
The action for the current timestep $a_t$ can be sampled from the categorical distribution. The most popular method for action sampling at inference-time is greedy search. Specifically, the most favorable action for current timestep $\hat{a}_t$ given the observed state $s_t$ is found by greedily selecting the most preferable action from the statistical distribution over the action space predicted by the actor:
\begin{equation}\label{eqn:greedy-rl}
    \hat{a}_t = \texttt{arg max}_{a}\pi_\theta (a, s_t) = \texttt{arg max}_{a\in \mathcal{A}}p_\theta (a | s_t).
\end{equation}
SGBS~\citep{neurips2022_39b9b60f} and RLGBS drawn an analogy between RL models and LLMs
showed that it is possible to apply beam search, a popular approach for natural language generation, to RL action generation, provided that the simulation environment is deterministic and that real-time performance is not necessary. Through beam search, the method sacrifices real-time performance, but allowed for parallel exploration of multiple action sequence hypotheses. At the end of the beam search procedure, favorable hypotheses, as characterized by high cumulative product of action-probabilities:
\begin{equation}\label{eqn:bs-rl}
    a_{1:T} = \texttt{arg max}_{a_t\in A}\prod_{t=1}^{T} p_\theta (a_t | a_{1:t-1}, s_{1:t}), a_{1:0} = \emptyset,
\end{equation}
can be evaluated and selected based on the final cumulative episodic reward. Compared to greedy search, SGBS and RLGBS often result in higher reward with a relatively efficient polynomial-time complexity in the exponential search space. The main difference between SGBS and RLGBS is that SGBS operates on problems with dense reward, while RLGBS operates on sparse reward problems ($r_t = 0$ until reaching the stopping condition). As a result, SGBS takes advantage of dense reward and prune beam candidates based on actual cumulative reward, while RLGBS relies purely on the beam score for beam candidate selection.

\subsubsection{Constrained Beam Search}
The earliest example of constrained beam serach was proposed in~\citep{anderson-etal-2017-guided} for image captioning, where a recurrent neural network (RNN) was tasked with generating short discriptive sentences for images. In order to ensure potentially unseen objects in the image are correctly included in the caption, the authors used a convolutional neural network (CNN) to perform object detection on the image before caption generation. The detected image tags are then enforced as \emph{disjunctive} constraints during the search. A \emph{disjunctive} constraint is a bundle of \emph{lexical} constraints that can be fulfilled by fulfilling just one of the member \emph{lexical} constraint. For example, the constraint $C_1 = \{\textrm{`desk'}, \textrm{`table'}\}$ is considered fulfilled when either `desk' or `table' appears in the generated text. Constraint enforcement is done by constructing a finite-state machine (FSM) that enumerates all possible paths in satisfying all constraints, and this results in a FSM whose number of states grows exponentially to the number of constraints. At generation time, a grouped beam search algorithm is used, where each group contains at most $n_b$ beams whose overall constraint fulfillment status corresponds to a state in the FSM. As beam candidates make progress in constraint fulfillment, they are moved to the corresponding beam group, and finally only beam candidates from the beam group that corresponds to ``all constraints fulfilled'' can be added to the finished hypotheses. Overall, the time complexity of the proposed CBS algorithm is $\mathcal{O}(T n_b 2^C)$, where $T$ is the length of the sentence, $n_b$ is the number of beams and $C$ is the number of constraints.

A faster alternative to CBS, named grid beam serach, was proposed in~\citep{hokamp-liu-2017-lexically} for machine translation problems. This method studies non-disjunctive \emph{lexical} constraints containing single words or phrases that must be fulfilled during text generation. This simplification of disjunctive constraints leads to a clearly defined ``length'' for each constraint. The authors constructed a grid with timestep $t$ as x-axis and constraint fulfillment status $c$ as y-axis to organize lexically constrained decoding. The grid beam search algorithm maintains $C+1$ banks/rows in the grid, with each bank containing up to $n_b$ beams. Every beam starts from the bottom-left of the grid and can either move up by opening/continuing a constraint, or move right by generating other tokens that do not contribute to constraint fulfillment. It should be noted that once a constraint is started, the beam is forced to continue it without the option of abandoning a constraint and possibly starting later. Finally, only beams that have reached the top bank of the grid (fulfilled all constraints) are added to the finished hypotheses. The time complexity of grid beam search is $\mathcal{O}(T n_b C)$.

Unlike conventional beam search, which maintains at most $n_b$ beams at any timestep, the actual number of beams maintained by CBS methods in~\citep{anderson-etal-2017-guided, hokamp-liu-2017-lexically} are either exponential or linear to the number of constraints, leading to poor scaling to growing number of constraints. In~\citep{post-vilar-2018-fast}, the authors introduced CBS through dynamic beam allocation (DBA) to further reduce the time complexity of the algorithm. Similar to conventional BS, DBA maintains up to $n_b$ beams at any timestep. Apart from top-$k$ candidates as selected based on the overall beam scores, DBA attempts to introduce additional candidates by adding tokens that would fulfill unmet constraints for the current beam candidates. The two groups of beam candidates are merged and grouped into banks based on their constraint fulfillment statuses. Finally, $n_b$ beams are divided into at most $C+1$ banks of approximately the same size and populated by the respective bank of beam candidates. The time complexity of DBA is $\mathcal{O}(T n_b)$. This method is further improved by Vectorized DBA~\citep{hu-etal-2019-improved} by using tries to keep track of disjunctive constraints and allowing for DBA to operate on GPU.

\subsection{Reinforcement Learning Constrained Beam Search}
As the name suggests, RLCBS is CBS applied to action generation with RL models. We adopted BS and CBS source code from the \texttt{GenerationMixin} module of Huggingface transformers~\citep{wolf-etal-2020-transformers}, an open source library for natural language processing with large language models. Huggingface transformer's CBS implementation is functionally equivalent to Vectorized DBA~\citep{hu-etal-2019-improved}. 

\begin{figure}
\centering
\includegraphics[width=\columnwidth]{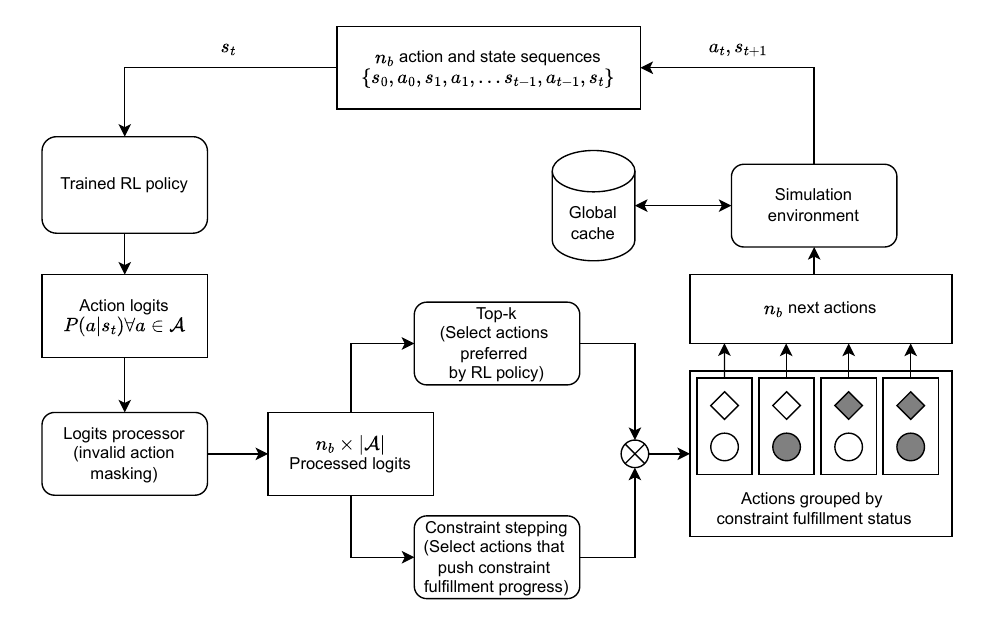}
\caption{Schematic showing one step in RLCBS. We start with $n_b$ beams, each represented by action-state sequence $\{s_0,a_0,... s_{t-1},a_{t-1},s_t\}$. For each beam, the RL policy predicts a discrete probability distribution across the action space $\mathcal{A}$, resulting in $n_b\times|\mathcal{A}|$ logits passed to the logits processor, where invalid actions are masked by setting the corresponding logits to $-\infty$. From the processed logits, two groups of actions are proposed: (1) top-$k$ actions as rated by the RL policy; and (2) actions that will advance the current beam in fulfilling the constraints. Subsequently, $n_b$ actions are selected from the proposed actions with high probabilities while maintaining a balanced mix of constraint fulfillment progress across the beam candidates. Finally, the next-actions $a_t$ are passed to the simulation environment to predict next state $s_{t+1}$ which will extend the beams by one action-state pair.}
\label{fig:rlcbs-algorithm}
\end{figure}

\subsubsection{RL Adaptation}
Figure~\ref{fig:rlcbs-algorithm} shows the schematic of the RLCBS algorithm. Although RLGBS is largely based on CBS for NLG, critical differences exist between action generation with an RL model and text generation with an LLM. While LLMs can generate text in a fully autoregressive manner, an RL agent relies on the simulation environment to predict the next states and continue the trajectory. For this purpose, a copy of the simulation environment serves as an oracle for sequence continuation and determining the stopping condition once the next-actions have been selected by CBS. During CBS, we do not use reward feedback from the oracle and rely solely on the RL policy output for selecting beams to keep. However, once a beam candidate has been confirmed completed by the oracle simulation environment and added to the completed hypotheses, the final beam hypothesis to return is selected based on actual cumulative episodic reward as calculated by the simulation environment.

\subsubsection{Negative Constraints}
Negative constraints (also referred to as preceding constraints in some literature) in RLCBS are handled by setting the desired action's log-probability (logit) to $-\infty$ in logits processors. In NLG, negative constraints are often used to exclude certain ``bad words'' or unwanted phrases from appearing in the generated text. In RL, negative constraints can be used to mask invalid actions or to set an upper limit of the number of times a certain class of action can appear in the generated sequence. The logits processor can be considered a more versatile solution to invalid action masking in Maskable~PPO~\citep{tang2020implementing}, as apart from masking unwanted actions, it is also possible to alter the action distribution to promote certain actions similar to guided beam search in NLG. We also use logits processors to renormalize logits after applying any negative constraint so that the probabilities of unmasked tokens sum to 1 after setting the invalid action logits to $-\infty$.

\subsubsection{Positive Constraints}
Examples of positive constraints include ``at least $n$ actions of certain type must appear in the generated sequence'' or ``exactly $n$ actions of certain type must appear in the generated sequence''. This type of constraints is difficult to reliably enforce through logits manipulation, as it is unclear at what stage of the episode the constraint should be fulfilled, and na\"ively pushing progress of constraints at arbitrary stages of an episode could lead to suboptimal constraint placement and consequently low cumulative episodic reward. In RLCBS, each positive constraint is encoded in a beam constraint object whose abstracted definition is provided in Algorithm~\ref{algo:abc-constraints}. For each beam candidate, the constraint object provides a list of actions that advances constraint fulfillment, processes the newly generated action token, and provides the steps remaining to complete the constraint. This information is required by the VDBA algorithm to correctly allocate banks across the beam candidate population. In the case of a disjunctive constraint with member constraints of different lengths, i.e. $C_1 = \{\textrm{`desk'}, \textrm{`coffee table'}\}$, the remaining length of the constraint is defined as the height of the disjunctive trie.

Constraints that specify the exact number of actions, such as ``exactly $n$ actions of a certain type that must appear in the generated sequence'', can be enforced by adding a beam constraint and a logit processor to RLCBS at the same time.

\begin{algorithm}
\begin{algorithmic}[1]
\Class{Constraint}
    \Function{advance}{}
        \State \Return list of tokens that advances constraint fulfillment
    \EndFunction
    \State
    \Function{update}{\texttt{token}}
        \State $\texttt{stepped} \gets \texttt{False}, \texttt{completed} \gets \texttt{False}, \texttt{reset} \gets \texttt{False}$
        \If{\texttt{token} creates progress}
            \State Progress constraint state
            \State $\texttt{stepped} \gets \texttt{True}$
            \If{Constraint completely fulfilled}
                \State $\texttt{completed} \gets \texttt{True}$
            \EndIf
        \ElsIf{constraint is phrasal}
            \State Reset constraint to initial state
            \State $\texttt{reset} \gets \texttt{True}$
        \EndIf
        \State \Return \texttt{stepped}, \texttt{completed}, \texttt{reset}
    \EndFunction
    \State
    \Function{remaining}{}
        \State \Return number of steps to complete constraint
    \EndFunction
\EndClass
\caption{Abstracted Constraint Class for Constrained Beam Search}\label{algo:abc-constraints}
\end{algorithmic}
\end{algorithm}

\subsubsection{Simulation Environment Caching}
Due to the size and computational cost of modern LLMs, the speed of BS/CBS text generation is often dominated by the speed of LLM inference. Conversely, the inference costs of RL models are often much lower due to their significantly lower number of parameters, and in the case of a simulation environment backed by an underlying physics-based model, the cost of simulating state transitions in the simulation environment can become dominant. For the sake of argument, consider the action-state sequence $\{s_0,a_0,... s_{t-1},a_{t-1},s_t\}$ with stopping condition $t = T$. To complete the sequence using conventional greedy search, a total of $T$ timesteps need to be simulated by the environment, as only one path from $s_0$ to $s_{\textrm{fin}}$ is explored. For CBS, since $n_b$ beams are considered in parallel, the simulation environment must restart $n_b$ times at each timestep to consider all beam candidates, leading to a total of $\frac{1}{2}T(T-1)n_b = \mathcal{O}(T^2 n_b)$ simulation steps before reaching the timestep $t=T$. To address the exponential growth in computational cost of physics-based simulation during CBS, we introduce a global cache that serves multiple processes running the simulation environments in parallel. The simulation environment's underlying C++ physics-based model can be serialized as a state vector encoding all internal states of the environment. During CBS, we employ the Algorithm~\ref{algo:redis-cache} where each simulation environment instance attempts to query the global key-value storage with the current action sequence as query key. In case of a cache miss, shorter prefixes of the action sequence will be used until a cache hit or the prefix becomes empty. The simulation environment then continues the simulation from the longest cached prefix, storing the state vectors obtained from the physics-based model after every simulation timestep. In single-thread execution, this scheme reduces the total number of simulation steps from $\frac{1}{2}T(T-1)n_b$ to $T n_b$. When multiple simulation environments are running in parallel, the total number of simulation timesteps may be higher than $T n_b$ due to multiple instances performing possibly duplicated work, but overall progress will be significantly faster due to parallelization. 

\begin{algorithm}
\begin{algorithmic}[1]
\Procedure{CachedSimulationRollout}{initial condition $s_{\textrm{init}}$, input IDs vector $\vec{a} = [a_1, a_2, \dots a_n]$}
    \For{$i \gets |\vec{a}|$ \textbf{down to} $1$}
        \If{$i = 1$}
            \State $\texttt{env} \gets \text{create\_environment}()$ \Comment{Base case: create new environment and simulate one step}
            \State $s_0 \gets \texttt{env}.\text{reset}(s_{\textrm{init}})$
            \For{$a \in \vec{a}[0:i]$}
                \State $s, r, \texttt{done} \gets \texttt{env}.\text{step}(a)$
            \EndFor
            \State \textbf{break}
        \EndIf
        \State $\textrm{key} \gets \text{JsonEncode}(s_{\textrm{init}}, \vec{a}[0:i])$
        \State $\texttt{value} \gets \text{cache}.\text{get}(\texttt{key})$ \Comment{If prefix exists in cache, retrieve environment state from cache}
        \If{$\texttt{value} \neq \emptyset$}
            \State $\texttt{env} \gets \text{create\_environment}()$
            \State $\texttt{env}.\text{set\_state}(\texttt{value})$ \Comment{Set new environment state to cache after every step}
            \State \textbf{break}
        \EndIf
    \EndFor
    \State $j \gets 0$
    \For{$a \in \vec{a}[i:]$} \Comment{Simulate any new actions after retrieving cached environment}
        \State $j \gets j + 1$
        \State $s, r, \texttt{done} \gets \texttt{env}.\text{step}(a)$
        \State $\texttt{key} \gets \text{JsonEncode}(s_{\textrm{init}}, \vec{a}[0:i+j])$
        \State $\texttt{value} \gets \texttt{env}.\text{get\_state}()$
        \State $\text{cache}.\text{set}(\texttt{key}, \texttt{value})$ \Comment{Set new environment state to cache after every step}
    \EndFor
    \State $s, r, \texttt{done} \gets \texttt{env}.\text{status}()$
    \State \Return $s, r, \texttt{done}$ \Comment{Return observation, reward, termination/truncation status to RL agent}
\EndProcedure
\caption{Simulation Environment Cached Rollout Function}\label{algo:redis-cache}
\end{algorithmic}
\end{algorithm}

\section{Smart Dryer Simulation Environment}
\label{sec:case_study}
\begin{figure}
\centering
	\includegraphics[width=\columnwidth]{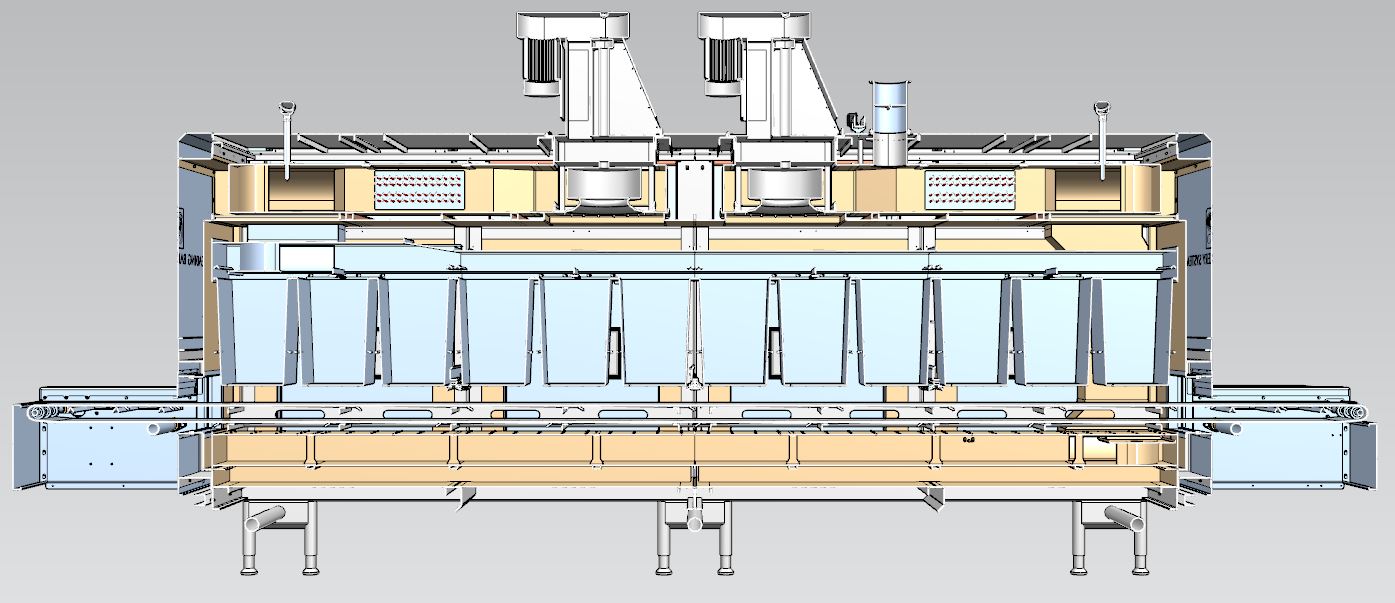}
	\caption{Section view of the Smart Dryer testbed. The testbed features a reconfigurable drying chamber that accommodates up to 12 dryer modules in a row. Paper samples are placed on a stainless steel mesh-type conveyer belt moving through the chamber at a constant set speed. Each of the 12 dryer positions can be occupied by one of the 4 dryer modules: slot-jet reattachment (SJR) module, perforated plate (PP) module, di-electrophoresis (DEP) module, and solid plate (SP) module. Finally, three IR emitters are installed between dryers 3-4, 6-7, and 9-10, and can be optionally enabled to assist in drying.}
	\label{fig:smart_dryer}
\end{figure}

This study considers a modular Smart Dryer testbed, shown in Figure~\ref{fig:smart_dryer}, which incorporates reconfigurable drying mechanisms to investigate the drying processes of pulp, paper, and other materials. The Smart Dryer is approximately \SI{9}{\meter} long with a drying chamber of length \SI{6.34}{\meter} (20.8 ft). One of the contributions of this work is the Smart Dryer drying simulation environment that facilitates training and evaluation of RL agents. The Smart Dryer simulation environment enables realistic and efficient data generation under randomized operating conditions, allowing for training on-policy RL algorithms with a reasonable computational budget. In this section, we describe the underlying theoretical drying model, the range of operating conditions, and a method for estimating overall energy consumption, which serves as the metric for evaluating the performance of the trained RL agent. A list of symbols referenced hereafter is given at the end of the paper.

\subsection{Physics-Based Drying Model}
To simulate the effects of various process parameters on the paper while it travels through the dryer, the Smart Dryer simulator leverages a theoretical drying model first proposed in~\citep{10.1115/1.2911313} for simulating free-water removal from paper in a conventional steam-heated, multi-cylinder dryer. The model is 1-D in space (thickness), transient, and predicts moisture and temperature profiles inside the paper while it travels through the series of dryer drums. Pilot machine trials verified the model on a multi-cylinder dryer incorporated with infrared emitters~\citep{seyed2001heating}.

Based on the theoretical model proposed in~\citep{10.1115/1.2911313}, we constructed a simulation model for the Smart Dryer and performed preliminary experiment verification, which shows good agreement between simulation and experiment results, as reported at the end of this section. 

The drying model is based on mass and energy conservation governing equations in the thickness direction:
\begin{equation}\label{eqn:mass_gov}
\frac{\partial M}{\partial t} = -\frac{\partial J_w}{\partial z} - \frac{\partial J_v}{\partial z}
\end{equation}
\begin{equation}\label{eqn:energy_gov}
(c_w \rho_w s \epsilon + c_f \rho_f (1-\epsilon)) \frac{\partial T}{\partial t} = - \frac{\partial q''}{\partial z} - \frac{\partial [J_wH_w+J_vH_v]}{\partial z} + \frac{\partial q''_{\textrm{IR}}|_z}{\partial z},
\end{equation}
Notably, the energy conservation equation has included the term ($\frac{\partial q''_{\textrm{IR}}|_z}{\partial z}$) as a volumetric internal heat source to represent the infrared energy penetration from the IR emitter modules~\citep{10.1115/1.1372324}.

Heat flux, liquid flux, and vapor fluxes are calculated by Fourier's, Darcy's, and Fick's Laws, respectively. 
\begin{equation}\label{eqn:fouriers_law}
q'' = -k_{\textrm{eff}} \frac{\partial T}{\partial z}
\end{equation}
\begin{equation}\label{eqn:darcys_law}
J_w = \left(\frac{K}{\nu_w}\right) \frac{\partial P_{ca}}{\partial z}
\end{equation}
\begin{equation}\label{eqn:ficks_law}
J_v = -\left(\frac{D_{ap}\textrm{MW}_v}{(1-y_v)}\right) \frac{\partial C_v}{\partial z}
\end{equation}

The governing Equations~(\ref{eqn:mass_gov}, \ref{eqn:energy_gov}) and boundary conditions for each drying phase are solved numerically using finite difference methods with the explicit Euler time-stepping method. The order of accuracy is first-order in time and second-order in space. The model uses 24 nodes in the paper thickness direction and a timestep size of \SI{0.025}{\ms}.

\begin{table}
\centering
\caption{Main parameters considered for optimization}
\label{tab:operating_conditions}
\begin{tabular}{ll}
\hline
\multicolumn{2}{l}{\textbf{Initial conditions, uncontrolled parameters}} \\ \hline
\multicolumn{1}{l}{Total drying time} & \SIrange{1.6}{16}{\minute} \\ 
\multicolumn{1}{l}{Machine speed $v_m$} & \SIrange{0.0066}{0.066}{\meter \per \second} \\ 
\multicolumn{1}{l}{Initial paper temperature $T_{p,\textrm{init}}$} & \SI{20}{\celsius} \\ 
\multicolumn{1}{l}{Initial paper $\textrm{DBMC}_{p,\textrm{init}}$} & 1.5 \\ 
\multicolumn{1}{l}{Target paper $\textrm{DBMC}_{p,\textrm{tgt}}$} & 0.2 \\ \hline
\multicolumn{2}{l}{\textbf{Controlled/optimized parameters}} \\ \hline
\multicolumn{1}{l}{Air temperature $T_a$} & $\SIrange{80}{232}{\celsius}$ \\ 
\multicolumn{1}{l}{Volumetric flow rate $\dot{V}_a$} & 200 to 500 SCFM \\ 
\multicolumn{1}{l}{IR temperature $T_{\textrm{IR}}$} & $\SIrange{427}{502}{\celsius}$ (IR1), $\SIrange{427}{677}{\celsius}$ (IR2), $\SIrange{427}{760}{\celsius}$ (IR3) \\ \hline
\end{tabular}%
\end{table}

The main process control parameters and operating condition ranges are presented in Table~\ref{tab:operating_conditions}. The physics-based model uses a fixed stopping condition of target dry-basis moisture content $\textrm{DBMC} \leq 0.2$, regardless of the initial condition. This is because the drying model primarily models the removal of free water in the paper, and the effects of bound water removal from within fibers become significant when the paper is dried to below 10\% DBMC~\citep{alma99955068842405899}. Given the random nature of the process operating conditions, keeping the drying target sufficiently above the threshold of 10\% DBMC is necessary to ensure that the simulation environment remains stable and outputs valid results even under aggressive process control parameter settings.

To facilitate RL training, a simulation environment wrapper was built after the OpenAI Gym/Gymnasium~\citep{openai-gym-1606.01540, towers_gymnasium_2023} base environment with measures to optimize the reliability and efficiency of the simulation model. The simulation environment employs smaller timestep sizes to ensure stability under highly variable process operating conditions, and key physics properties of the material are continuously monitored during simulation. For example, the paper temperature is ensured not to exceed boiling temperature, and any invalid results will lead to the session being truncated prematurely. The simulation environment was written in C++ to optimize computational efficiency while allowing for interfacing with the RL agent through \texttt{pybind11} bindings. The dryer simulation environment simulates a 4-minute drying experiment in approximately \SI{40}{\s} on a single CPU thread, and additional speed improvements are possible through parallelization across multiple threads or computing nodes. We estimate that the compiled C++ simulation program is approximately 140 times faster than the Python prototype before accounting for effects of parallelization. We have additionally designed the internal state of the simulation program to be serializable, and implement a global cache to eliminate duplicate computation loads when the process initial conditions are fixed, such as during RLCBS action generation process.

\subsubsection{Boundary Conditions}
One of the distinctive features of the Smart Dryer is the reconfigurable drying chamber which accommodates up to 12 dryer modules in a row. Each of the 12 dryer positions can be occupied by one of the 4 dryer modules: slot jet reattachment (SJR) module, perforated plate (PP) module, dielectrophoresis (DEP) module, and solid plate (SP) module. Among these, the SJR and PP modules receive hot air supply with temperature $T_{a}$ and velocity $v_{a}$, while the DEP and SP modules rely on the natural convection effect for drying, as the hot air supply is blocked at these locations. Finally, three IR emitters are installed between dryers 3-4, 6-7, and 9-10 and can optionally be enabled to assist in drying. As the paper progresses in the dryer, the effects of each drying module on moisture removal from the paper sheet are reflected in the boundary conditions imposed on top and bottom surfaces of the paper that correspond to the respective dryer module's behavior.

Compared to~\citep{10.1115/1.2906435} the boundary conditions are adjusted as the porous sheet is no longer in contact with a drying cylinder under contact pressure, requiring new boundary conditions for the bottom side ($z=0$) and for the top side ($z=Z$) that describe the mass and heat transfer to be formulated.

On the bottom surface grid, the paper is in contact with the dryer for boundary conditions~(\ref{eqn:mass_bc_bottom}, \ref{eqn:energy_bc_bottom}). As there is no tension applied on the paper handsheet placed on the conveyer belt, the contact pressure-based dryer-paper web interface thermal contact conductance $h_{i}$ will be significantly lower following the correlation in~\citep{10.1115/1.2906435}. Conversely, an adiabatic boundary condition is assumed for the bottom surface, since the paper is sitting on a thin metal plate that has a low Biot number ($\textrm{Bi} \ll 0.1$), then $T_{su}-T=0$ leaving the interface contact conductance ($h_{i}$) unused and again omit the term $h_{i}(T_{su}-T)$ in the simulation.
\begin{equation}\label{eqn:mass_bc_bottom}
\frac{\partial M}{\partial t} = \frac{1}{\Delta z}(-J_w - J_v)
\end{equation}
\begin{equation}\label{eqn:energy_bc_bottom}
\begin{split}
&(c_w \rho_w s \epsilon + c_f \rho_f (1-\epsilon)) \frac{\partial T}{\partial t} \\
&= \frac{1}{\Delta z}(h_{i}(T_{su}-T) + k_{\textrm{eff}} \frac{\partial T}{\partial z} - J_wH_w - J_vH_v + q''_{\textrm{IR}}|_{z=\Delta z} - q''_{\textrm{IR}}|_{z=0})
\end{split}
\end{equation}

On the top surface grid, paper is exposed to air. The boundary conditions~(\ref{eqn:mass_bc_top}, \ref{eqn:energy_bc_top}) contains the term $-h(T-T_{a})$ that accounts for the convective heat flux $q_{conv}$.
\begin{equation}\label{eqn:mass_bc_top}
\frac{\partial M}{\partial t} = \frac{1}{\Delta z}(J_w + J_{v,i} - J_{v,o})
\end{equation}
\begin{equation}\label{eqn:energy_bc_top}
    \begin{split}
&(c_w \rho_w s \epsilon + c_f \rho_f (1-\epsilon)) \frac{\partial T}{\partial t} \\
&= \frac{1}{\Delta z}(- k_{\textrm{eff}} \frac{\partial T}{\partial z} + J_wH_w + J_{v,i}H_v - (1 + \textrm{DRE}_{\textrm{DEP}})J_{v,o}H_v - h(T-T_a) + q''_{\textrm{IR}}|_{z=Z} - q''_{\textrm{IR}}|_{z=Z-\Delta z})
    \end{split}
\end{equation}

The convective heat transfer coefficient $h$ in Equation~\ref{eqn:energy_bc_top}, determined through experimental correlations for each dryer module type, varies with the relative position $x$ of paper in the dryer, where the $h$ value represents the average measurement obtained along the entire dryer length during experimental trials.

The radiation heat flux term in the amount of energy that penetrates the paper in the thickness direction is exponentially decaying to follow Bouguer's law as in Equation~\ref{eqn:bouguers_law}, where $Z$ is the total sample thickness, $q''_{\textrm{IR}}|_{z=Z}$ represents the incident radiation on the top surface, and $a$ is absorption coefficient calculated based on that of fiber ($a_f$), water ($a_w$) weighted by current saturation. This term is only considered under the influence of the IR module, adjusted by its view factor dynamically according to the relative position~\citep{10.1115/1.1372324}.

\begin{equation}\label{eqn:bouguers_law}
    q''_{\textrm{IR}}|_{z} = q''_{\textrm{IR}}|_{z=Z} \exp\left(\int_{Z}^{z} a \, dz\right)
\end{equation}

The boundary conditions shown above have also accounted for different drying technologies, including internal heating caused by penetration of infrared energy and drying rate enhancement by DEP ($\textrm{DRE}_{\textrm{DEP}}$) in a general form. The DEP module enhances the vapor flux within the porous media and is not expected to influence the convective heat transfer. The correlation of $\textrm{DRE}_{\textrm{DEP}}$ with DBMC ($M_d$) in Equation~\ref{eqn:dep_corr} is curve-fitted~\citep{sarikaya2022fundamental} based on the experimental drying curve in~\citep{doi:10.1080/07373937.2021.1981922}. Such impacts are reflected when the paper is within the boundaries of a corresponding dryer module.

\begin{equation}\label{eqn:dep_corr}
\textrm{DRE}_{\textrm{DEP}} = (19133M_d^6-94421M_d^5+185596M_d^4-185349M_d^3 +99171M_d^2-27003M_d+2952.4)\times 100\%
\end{equation}

\subsection{Estimation of Energy Consumption}

To simulate the effects of various drying technologies and control parameters on the paper being dried, as well as provide a basis for optimization, a physics-based drying model was developed to numerically solve the paper temperature $T_{t}$ and dry-basis moisture content $\textrm{DBMC}_{t}$ values at any time $t$ during the drying process. From the simulated temperature and the DBMC profiles, the overall energy consumption $q$, in terms of kilojoules per square meter of dried paper (\SI{}{\kjmm}), could be estimated as in Equation~\ref{eqn:q_definition}:

\begin{equation}\label{eqn:q_definition}
    q = \sum_{t=0}^{t=t_{\textrm{fin}}}  (-\Delta\overline{\textrm{DBMC}}_{t:t-1} \cdot \textrm{BW} \cdot h_{fg} + \Delta\overline{T}_{t:t-1} \cdot \textrm{th}_t(c_w \rho_w \epsilon_t \overline{s}_t + c_f \rho_f (1 - \epsilon_t))),
\end{equation}
\noindent where $\Delta\overline{\textrm{DBMC}}_{t:t-1}$ and $\Delta\overline{T}_{t:t-1}$ denote the change in mean DBMC and temperature across the thickness direction of the paper over the last simulation timestep, respectively. The total system power can be estimated by assuming constant throughput (i.e., area of paper that is dried in the dryer at any given time). Although the estimated system power may not be accurate, determining control strategies that minimize $q$ in simulation could help to find strategies to implement energy-efficient control on the actual testbed.

\begin{figure}
\centering
\includegraphics[width=0.8\columnwidth]{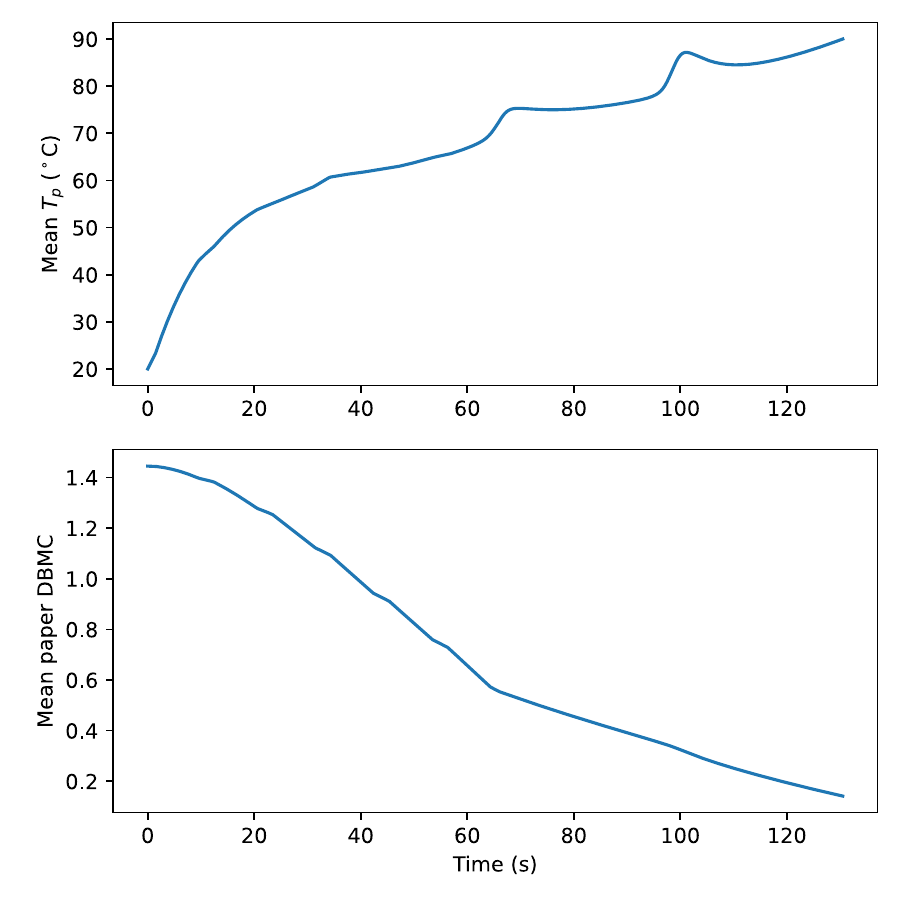} 
\caption{Sample paper temperature and dry-basis moisture content (DBMC) trajectory as simulated by the physics-based drying model using dryer configuration: 6x SJR, 6x SP and process parameters: total drying time \SI{2.2}{\minute}, machine speed $v_m = \SI{0.0482}{\meter \per \second}$, hot air supply temperature $T_a = \SI{140.20}{\celsius}$, air velocity $v_a = \SI{11.41}{\meter \per \second}$, IR temperature $T_{\textrm{IR}_1} = \SI{501.99}{\celsius}$, $T_{\textrm{IR}_2} = \SI{677.08}{\celsius}$, $T_{\textrm{IR}_3} = \SI{760.0}{\celsius}$.}
\label{fig:sample_trajectory}
\end{figure}

\subsection{Experimental Validation}

Experimental validation is conducted on the testbed under the specified conditions. The test samples are lab-made paper handsheets of refined hardwood, prepared following the TAPPI T 205 standard~\citep{tappi_t405cm20}. The test is repeated three times with freshly made handsheets at room temperature. The prepared samples have a basis weight of \SI{0.2388}{\kg \per \meter \squared} and an initial DBMC of 1.4447 ($\sigma = 3.47\%$). The drying process parameters are: total drying time \SI{2.2}{\minute}, machine speed $v_m = \SI{0.0482}{\meter \per \second}$, hot air supply temperature $T_a = \SI{140.20}{\celsius}$, air velocity $v_a = \SI{11.41}{\meter \per \second}$, IR temperature $T_{\textrm{IR}_1} = \SI{501.99}{\celsius}$, $T_{\textrm{IR}_2} = \SI{677.08}{\celsius}$, $T_{\textrm{IR}_3} = \SI{760.0}{\celsius}$. Under these conditions, the final measured DBMC of the dried samples is 0.1385 ($\sigma = 0.84\%$). Simulating the drying process with the same parameters yields a final DBMC of 0.1406, differing by 1.52\% from the experimental result. Due to the final DBMC target being lower than our conventional value of 0.2, we reduced the simulation timestep size from \SI{0.25}{\ms} (40,000 timesteps per second) to \SI{0.154}{\ms} (65,000 timesteps per second) to ensure that the simulation completes without error. The simulated temperature and DBMC profiles for these conditions are presented in Figure~\ref{fig:sample_trajectory}.

The measured gross power input to the dryer is \SI{44.56}{\kilo \watt}, which includes all installed modules, air heating, blowers, and conveyor belt. For a fully loaded belt, this translates to \SI{12.0}{\kilo \watt \per \meter \squared}. The drying rate of the validation samples is calculated at \SI{2.36}{\gram \per \meter \squared \per \second}, equivalent to \SI{5.31}{\kilo \watt \per \meter \squared} for a continuously fed paper-covered belt. The energy efficiency ratio is 44\%, with potential for significant improvement through exhaust energy recovery commonly found in an industrial setup.

\section{Experiment Setup}
\label{sec:experiment_setup}
We approach the problem of combinatorial optimization of dryer module assignment and hot air supply temperature in the Smart Dryer in a two-step process. The problem is first formulated as a Markov decision process without constraints and solved using RL techniques. Then, we define design constraints in RLCBS and apply the algorithm to RL action generation at inference time to decode high-quality action sequences that maximize cumulative reward while respecting the inference-time constraints. The RL policy trained in the unconstrained environment is used to guide exploration in the exponential search space while bound by negative and positive constraints.

\subsection{RL Model Training}
The Smart Dryer is discretized into up to 12 timesteps of equal lengths, each corresponding to the amount of time required for the paper to travel through the span of a drying module. For the RL task, it is assumed that dryer modules can be freely reconfigured with hot air at potentially varied temperature levels supplied to each module. Although varied temperature air supply for different modules is currently exceeding the capacity of the physical dryer, the intention is that the RL model will be trained in an environment with minimal constraints and additional constraints can be programmed into RLCBS at inference-time if needed. 

During RL training and inference, the simulation is carried out such that each dryer module represents a timestep. At each timestep $t$, the RL policy $\pi$ observes the state of the environment $s_t$ and chooses an action $a_t$ from the finite action space $\mathcal{A}$, which specifies the module type and air supply tempearture to be used in the next timestep. The environment proceeds by one timestep upon receiving the action input and provides the RL agent with the next state $s_{t+1}$ and a scalar reward $r_t$ as the feedback. The interaction between the RL agent and the environment repeats until a stopping condition is reached at timestep $T$. The RL agent is trained to find the optimal policy that maximizes the cumulative discounted reward for all future timesteps $t$ to $t_{fin}$:
\begin{equation}\label{eqn:cumul-reward-definition}
R_t = \sum_{\tau=t}^{T} \gamma^{\tau-t} r_\tau,
\end{equation}
through actions $a_{t:T}$ supplied to the environment.

Proximal Policy Optimization (PPO)~\citep{schulman2017proximal}, a popular on-policy algorithm, is used to learn the optimal RL policy. It falls into the category of actor-critic RL algorithms where a policy (actor) and a value function (critic) are jointly trained by interacting with the environment $\mathcal{E}$ over a series of discrete timesteps.

The actor is a neural network policy $\pi_\theta(s_t)$ parameterized by $\theta$. It maps the state space $\mathcal{S}$ to a categorical distribution over the discrete action space $\mathcal{A}$. Specifically, the actor network accepts a state vector containing temperature and DBMC at the top and bottom surface of the paper and machine speed. The output from the actor network is a multi-discrete distribution, comprised of two categorical distributions covering the 4 module types and 11 discrete temperature levels equally displaced across the defined hot air temperate range of 80 to \SI{190}{\celsius}\footnote{The upper limit of \SI{190}{\celsius} is lower than actual capabilities of the physical dryer at \SI{232}{\celsius}, however, we found that $T_a$ up to \SI{190}{\celsius} is already sufficient for drying our samples at total drying time of \SI{2}{\minute} or above, and excessive $T_a$ frequently leads to failure of the physics-based model.}. The multidiscrete output setup simplifies the combinatorial action space and allows for easy scaling to denser $T_a$ grades in the future. Final action output is formed by applying outer product between the module state vector and air temperature state vector, and reshaping the result into a $n \times 44$ array where $n$ is batch size:
\begin{equation}
\pi_\theta(s_t) = \{p_\theta(a|s_t)  \forall a \in \mathcal{A}\}.
\end{equation}

The action for current timestep $a_t$ can be sampled from the categorical distribution. We present the details for the state and action variables in Table~\ref{table:states-actions}.

\begin{table}
\centering
\caption{Range of state and action state variables expected by the PPO RL agent. State space variables are normalized to 0-1 using the running mean and variance during training. The RL agent outputs a multi-discrete distribution that is post-processed to form the combined action space before the actions are sampled and passed to the simulation environment.}
\label{table:states-actions}
\vspace{6pt}
\resizebox{0.7\columnwidth}{!}{%
\begin{tabular}{ll}
\hline
\textbf{Name} & \textbf{Range} \\ \hline
\textbf{State space} &  \\
Speed factor & 0.25-0.75 (actual machine speed: 0.0215-\SI{0.0512}{\meter \per \second}) \\
Mean paper temperature & 20-\SI{100}{\celsius} \\
Mean paper DBMC & 0.0-1.5 \\ 
Current position & 0-12 dryers (relative to beginning of drying process) \\ \hline
\textbf{Action space} &  \\
Dryer module type & 0 (PP), 1 (SJR), 2 (DEP), 3 (SP) \\
Hot air temperature $T_a$ & $\left[ 80,  91, 102, 113, 124, 135, 146, 157, 168, 179, 190\right]\SI{}{\celsius}$ \\ 
Combined action space & 0-10 (PP), 11-21 (SJR), 22-32 (DEP), 33-43 (SP)\\ \hline
\end{tabular}%
}
\end{table}

The critic is a neural network value function $V_\phi(s_t)$ parameterized by $\phi$ that, similar to the actor, takes as input the current state of the paper and outputs an estimated future return:
\begin{equation}
    V_\phi(s_t) = \hat{\mathbb{E}}[R_t|s_t, \pi_\theta].
\end{equation}

We use an implementation of the PPO~\citep{schulman2017proximal} algorithm from Stable-baselines3~\citep{stable-baselines3} to train the RL agent, which uses a fully connected multilayer perceptron (MLP) for both the policy (actor) and the value function (critic). More details on the structure and hyperparameters of the RL network can be found in Table~\ref{tab:ppo-hyperparams}. At training time, we split the workload into 30 processes, each maintaining a copy of the RL agent and a simulation environment with a unique random seed for randomizing process operating conditions. Before being passed to the RL model, the state and reward variables are individually normalized using the running mean and standard deviation calculated from past observations during training. We keep the best checkpoints at set timestep intervals, identified by the highest mean cumulative episodic reward over the last 100 episodes. At the end of training, we save the then-best checkpoint as the final model weights. 

\begin{table}
\centering
\caption{Network structure and hyperparameters used in training of the PPO RL agent with multi-discrete action space.}
\vspace{6pt}
\label{tab:ppo-hyperparams}
\renewcommand{\arraystretch}{0.9}
\resizebox{0.7\columnwidth}{!}{%
\begin{tabular}{lllll}
\hline
\multicolumn{4}{l}{\textbf{Network Structure}} \\
 & Input features & Output features & Bias & Activation \\ \hline
\multirow{3}{*}{Policy net} & $|\mathcal{S}|=6$ & 64 & \multirow{3}{*}{yes} & \multirow{2}{*}{tanh} \\
 & 64 & 64 & & \\
 & 64 & 15 (11 + 4) & & softmax \\ \hline
\multirow{3}{*}{Value net} & $|\mathcal{S}|=6$ & 64 & \multirow{3}{*}{yes} & \multirow{3}{*}{tanh} \\
 & 64 & 64 & & \\
 & 64 & 1 & &  \\ \hline
\multicolumn{5}{l}{\textbf{Hyperparameters}} \\
Learning rate & $3\times 10^{-4}$ & Timesteps per update & 2048 \\
Num. CPUs & 30 & Max. timesteps & $1\times 10^7$ \\
$\lambda$ (GAE) & 0.95 & $\gamma$ (reward discounting) & 0.99 \\
$\epsilon$ (clip factor) & 0.2 & Max. KL divergence & 0.01 \\ \hline
\end{tabular}%
}
\end{table}

\subsubsection{Reward Function}
The main objective of training the RL agent is to minimize overall energy consumption, regardless of the machine speed setting set at the beginning of each episode. However, cumulative energy consumption $q$, defined in Equation~(\ref{eqn:q_definition}), cannot be used directly for feedback to the RL agent, as it depends closely on the machine speed, with faster drying leading to higher energy consumption even in optimal settings. To ensure consistent feedback that encourages minimizing energy consumption in the face of variable initial conditions, we form a reward function by comparing the RL drying outcome with a pre-calculated baseline:
\begin{equation}\label{eqn:reward-function}
r_t = \begin{cases}
q_{\textrm{SQP}} - \sum_{i=1:t} q_i &\text{$t \leq T_{\textrm{max}}$ \textrm{and} $\textrm{DBMC}_{t} \leq \textrm{DBMC}_{\textrm{tgt}}$}\\
q_{\textrm{SQP}} - \sum_{i=1:t} q_i - 1000 &\text{$t = T_{\textrm{max}}$ \textrm{and} $\textrm{DBMC}_{t} > \textrm{DBMC}_{\textrm{tgt}}$}\\
0 &\text{otherwise},
\end{cases}
\end{equation}
where $q_t$ is the energy consumption during timestep $t$, $T_{\textrm{max}} = 12$ is the maximum number of drying modules allowed by the physical Smart Dryer testbed, and $q_{\textrm{SQP}}$ is a cumulative energy consumption baseline found with sequential quadratic programming (SQP)~\citep{wilson1963_simplicial} by optimizing global process variables $T_a$, $v_a$ under fixed dryer module configuration for the current machine speed level. Essentially, $r_t$ is a sparse reward function that is nonzero if and only if one or more of the stop conditions are met: (1) paper DBMC is reduced to below target $\textrm{DBMC}_{\textrm{tgt}} = 0.2$; and/or (2) paper has traveled to the end of the 12th drying module. If (1) is fulfilled, the RL agent is rewarded by the energy saved compared to SQP baseline; otherwise, the RL agent is penalized by a negative reward.

\subsubsection{SQP Reward Baseline} SQP is a well-known iterative method for constrained nonlinear optimization. Here, $q_{\textrm{SQP}}$ represents the lowest possible overall process energy consumption achieved over a predefined sequence of 12 dryer modules and constant global process parameters, i.e., all dryer modules share the same hot air temperature $T_a$, flow rate $v_a$, IR on/off status and IR surface temperature. 

Using SQP, we attempt to solve the following constrained nonliear optimization problem:
\begin{equation}\label{eqn:sqp-problem-definition}
\begin{aligned}
\text{Minimize:} \quad & q = f(T_a, v_a, T_{\textrm{IR}}, T_{p,\textrm{init}},  \textrm{DBMC}_{p,\textrm{init}}, \textrm{SF}), \\
\text{Subject to:} \quad & \textrm{DBMC}_{p,\textrm{fin}} - 0.2 \leq 0, \\
& \textrm{DBMC}_{p,\textrm{fin}} = g(T_a, v_a, T_{\textrm{IR}}, T_{p,\textrm{init}}, \textrm{DBMC}_{p,\textrm{init}}, \textrm{SF}), \\
\text{With:} \quad & \SI{80}{\celsius} \leq T_a \leq \SI{232}{\celsius}, \quad \SI{427}{\celsius} \leq T_{\textrm{IR}} \leq \SI{760}{\celsius}, \\
& v_a = \frac{\dot{V}_a}{\rho_a A_{\textrm{nozzle}}}, \quad 200 \textrm{ SCFM} \leq \dot{V}_a \leq 500 \textrm{ SCFM},\\
& T_{p,\textrm{init}} = \SI{20}{\celsius}, \quad \textrm{DBMC}_{p,\textrm{init}} = 1.5, \quad A_{\textrm{nozzle}} = \SI{0.0233}{\meter \squared}
\end{aligned}
\end{equation}

Here, $f$ and $g$ are objective and constraint functions backed by the simulation environment. $T_a$, $T_{\textrm{IR}}$ are normalized to $[0, 1]$ before optimization. Air velocity $v_a$ is determined collectively by fan power (air volumetric flow rate) and $\rho_a$, which in turn is influenced by $T_a$. Machine speed, $v_{\textrm{m}}$, controls total drying time and is manually specified for each optimization run by specifying speed factor, $\textrm{SF}$. Conversion between $v_{\textrm{m}}$ and speed factor $\textrm{SF}$ is done as follows:
\begin{equation}\label{eqn:speed-factor-vs-machine-speed}
v_{\textrm{m}} = v_{\textrm{m,min}} + (v_{\textrm{m,max}} - v_{\textrm{m,min}}) \times (1 - \textrm{SF}),
\end{equation}
where $v_{\textrm{m,min}} = \SI{0.006604}{\meter \per \second}$ corresponds to theoretical maximum drying time of \SI{16}{\minute} and $v_{\textrm{m,max}} = \SI{0.06604}{\meter \per \second}$ corresponds to theoretical minimum drying time of \SI{1.6}{\minute}. Finally, the dryer sequence configuration is fixed at 6 SJR modules followed by 6 PP modules. We perform two sets of experiments where all IR modules are enabled in the first and disabled in the second. 

\begin{figure}
\centering
\includegraphics[width=0.5\columnwidth]{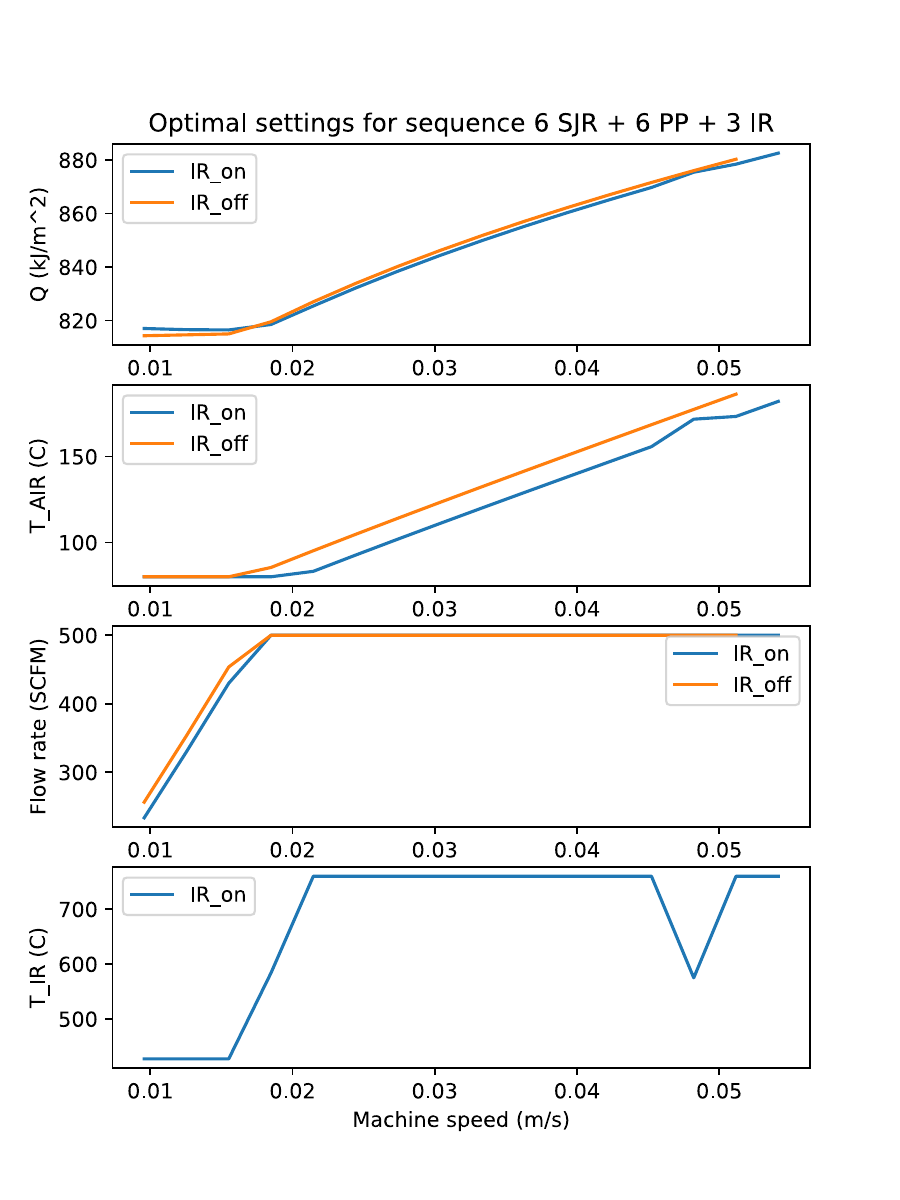}
\caption{Optimized air temperature $T_a$, air flow rate (in SCFM), IR surface temperature (assuming $T_{\textrm{IR}_1} = T_{\textrm{IR}_2} = T_{\textrm{IR}_3}$), and the resulting overall energy consumption $q$ as functions of machine speed given fixed dryer sequence configuration: SJR in positions 1-6, PP in positions 7-12, and IR on/off between dryers 3-4, 6-7, and 9-10.}
\label{fig:sqp_results}
\end{figure}

We use a SQP implementation provided in MATLAB r2024a. Communication between MATLAB and the C++ physics-based model is through an interface generated using MATLAB \texttt{clibPublishInterfaceWorkflow}. The baseline results optimized by SQP are shown in Figure~\ref{fig:sqp_results}. The results are omitted at machine speed levels where SQP cannot find a feasible solution. The reason for failing to find an feasible solution could include (1) failing to satisfy constraint $\textrm{DBMC}_{p,\textrm{fin}} - 0.2 \leq 0$ even when all controlled variables are maximized, which occurs when the machine speed is too fast; or (2) physics-based model fail to complete simulation under the prescribed set of controlled variables, which is most likely to occur when machine speed is too slow. We found that SQP is very sample efficient in this problem, requiring on average less than 30 objective function calls to converge for each machine speed level. However, since SQP function calls must be made sequentially and cannot be fully parallelized, each session requires approximately \SI{1}{\hour} wall time even with central differencing parallelization enabled.

From the SQP optimized baseline, we note the following observations.
\begin{itemize}
    \item Compared to disabling IR, Enabling IR results in higher maximum feasible machine speed, but does result in less than \SI{3}{\kjmm} energy savings on average. We conclude that IR emitters increase the maximum drying capacity of the dryer, but does not result in significant energy savings.
    \item When $\textrm{SF} > 0.75$ or $v_{\textrm{m}} < \SI{0.0215}{\meter \per \second}$, the optimal strategy is to minimize $T_a$ and $T_{\textrm{IR}}$ and reduce $\dot{V}_a$ to avoid overdrying. Below $v_{\textrm{m}} \approx \SI{0.0176}{\meter \per \second}$ or the total drying time over \SI{6}{\minute}, overdrying becomes inevitable.
    \item Between $0.25 \leq \textrm{SF} \leq 0.75$ or $\SI{0.0215}{\meter \per \second} \leq v_{\textrm{m}} \leq \SI{0.512}{\meter \per \second}$, the optimal policy is found by maximizing $\dot{V}_a$ and adjusting $T_a$ and $T_{\textrm{IR}}$ (if IR is enabled).
    \item With a total drying time less than \SI{2.06}{\minute}, or $\textrm{SF} < 0.25$, $v_{\textrm{m}} > \SI{0.512}{\meter \per \second}$, SQP often fails to find a feasible solution that satisfies the constraint $\textrm{DBMC}_{p,\textrm{fin}} - 0.2 \leq 0$, likely due to the drying of the specified sample exceeds the capacity of the dryer at these machine speed levels.
\end{itemize}

Based on the observations, the following steps are taken to arrive at the SQP baseline used in Equation~\ref{eqn:reward-function}:
\begin{itemize}
    \item We disable all IR emitters in the RL CO problem for simplicity. IR irradiation affects the entire drying chamber, making it challenging for the RL agent to control the IR status in an autoregressive manner. The only feasible option would be to enable IR at the beginning of each episode similarly to machine speed, but this would add unnecessary complexity to the problem. The simulation model was also found to be more stable with IR disabled.
    \item We constrain the machine speed range for RL CO problem to $0.25 \leq \textrm{SF} \leq 0.75$ or $\SI{0.0215}{\meter \per \second} \leq v_{\textrm{m}} \leq \SI{0.512}{\meter \per \second}$, while maintaining $v_a$ at its maximum value. This range is chosen because, outside it, the solution for $T_a$ becomes trivial. Operating within this speed range also eliminates the need for the RL agent to control $\dot{V}_a$, thus simplifying the problem.
\end{itemize}

The SQP baseline energy consumption values for $\SI{0.0215}{\meter \per \second} \leq v_{\textrm{m}} \leq \SI{0.512}{\meter \per \second}$ are shown in Table~\ref{tab:sqp-baseline-values}. For any unseen machine speed levels, linear interpolation is used to find an estimate for $q_{\textrm{SQP}}$. The reward function in Equation~(\ref{eqn:reward-function}) then directly measures RL energy savings compared to the baseline and encourages energy-saving action inputs by assigning positive rewards whenever the RL policy achieves lower energy consumption compared to the baseline and vice versa.

\begin{table}
\caption{SQP-optimized global process control parameters $T_a$, $\dot{V}_a$ (normalized) and resulting energy consumption  used in the RL reward function, assuming dryer sequence configurations are fixed, $T_a$ is constant for all drying modules, all IR modules disabled, and $v_a$ is set at maximum value.}
\vspace{6pt}
\label{tab:sqp-baseline-values}
\centering
\resizebox{\columnwidth}{!}{%
\begin{tabular}{lllll}
\hline
\textbf{Speed factor} & \textbf{Machine speed $v_{\textrm{m}}$ (\SI{}{\meter \per \second})} & \textbf{Normalized $T_a$} & \textbf{Normalized $\dot{V}_a$} & \textbf{Energy consumption (\SI{}{\kjmm})} \\
\hline
0.25 & 0.0512 & 0.9659 & 1 & 879.1134 \\
0.30 & 0.0482 & 0.8853 & 1 & 874.8082 \\
0.35 & 0.0452 & 0.8044 & 1 & 870.3372 \\
0.40 & 0.0422 & 0.7232 & 1 & 865.6814 \\
0.45 & 0.0393 & 0.6417 & 1 & 860.8234 \\
0.50 & 0.0363 & 0.5596 & 1 & 855.7368 \\
0.55 & 0.0334 & 0.4771 & 1 & 850.3909 \\
0.60 & 0.0304 & 0.3938 & 1 & 844.7547 \\
0.65 & 0.0274 & 0.3096 & 1 & 838.7919 \\
0.70 & 0.0244 & 0.2244 & 1 & 832.4452 \\
0.75 & 0.0215 & 0.1377 & 1 & 825.6498 \\
\hline
\end{tabular}
}
\end{table}

\subsection{RLCBS with Inference-Time Design Constraints}
We introduce three design constraints based on practical considerations of the physical drying testbed, encompassing both negative and positive constraints:
\begin{enumerate}
    \item \textbf{Force maximum 6 SJR modules:} The generated sequence must include no more than six SJR modules. Initial experiments showed that SJR modules have the highest drying capacity and are more efficient than PP modules, making them generally favored by the learned RL policy. However, SJR modules are more expensive to manufacture than PP modules. Additionally, due to their larger air outlet compared to PP modules, installing too many SJR modules in the dryer may result in uneven and insufficient airflow distribution to other modules.
    \item \textbf{Force minimum 3 DEP modules:} The generated sequence must include at least 3 DEP modules. Although DEP modules have relatively low drying capacity, they are highly energy efficient, consuming less than \SI{100}{\watt} per unit. Including more DEP modules in the solution sequence is desirable, even if they are not preferred by the learned RL policy.
    \item \textbf{DEP/SP air temperature continuity:} This constraint arises from the nature of DEP and SP modules: the dryer does not actively supply hot air to these positions, as air passage is blocked by the DEP module or by a solid plate. Consequently, the setting of air temperature ($T_a$) for the DEP and SP modules must match that of the preceding module in the solution sequence. 
\end{enumerate}

In RLCBS, negative constraints 1, 3 are implemented as logits processors, followed by logits renormalization. Specifically, constraint 1 masks any actions that correspond to an SJR module (actions 11-21) once there are already 6 or more such actions in the action sequence prefix. Constraint 3 masks any actions that correspond to a DEP or SP module that does not share the same air temperature setting with the preceding module, but this constraint will not be enabled in the first step of each episode due to the lack of a preceding module in such case.

Positive constraint 2 is implemented as a \texttt{SequentialDisjunctiveConstraint} that can be fulfilled by three actions that correspond to DEP modules (action 22-32) in the generated action sequence. Different from a \texttt{PhrasalConstraint}, the prescribed actions do not have to be adjacent in the sequence. Note that the SQP baseline for RL training uses a dryer module sequence configuration (6x SJR, 6x PP) that violates constraint 2. It is infeasible to train an RL agent for every combination of design constraints, as the reward function cannot be easily changed after training and each training session requires significant time investment due to the computational intensity of the physics-based simulation. Nevertheless, the SQP baseline should remain reflective of the overall trend of energy consumption versus machine speed even when some design constraints are violated.

\subsection{NSGA-II Baseline}
We use the Non-dominated Sorting Genetic Algorithm (NSGA-II)~\citep{996017} as a direct comparison of RLCBS in the task of simulataneously optimizing dryer sequence optimization and module-specific hot air supply temperature. We select NSGA-II as it supports all types of constraints discussed in this paper, and is widely recognized as a popular multi-objective evolutionary algorithm that achieves good performance in a variety of combinatorial optimization problems, allowing us to validate whether RLCBS has approached global optimum by comparing RLCBS and NSGA-II results under the same experiment configuration.

Using NSGA-II, we attempt to solve the following constrained combinatorial optimization problem:
\begin{equation}\label{eqn:nsga2-problem-definition}
\begin{aligned}
\text{Minimize:} \quad & q = f(T_{a,d=1:n}, v_a, \textrm{MT}_{d=1:n}, T_{p,\textrm{init}}, \textrm{DBMC}_{p,\textrm{init}}, \textrm{SF}), \\
\text{Subject to:} \quad & \textrm{DBMC}_{p,\textrm{fin}} - 0.2 \leq 0, \\
& \textrm{DBMC}_{p,\textrm{fin}} = g(T_{a,d=1:n}, v_a, \textrm{MT}_{d=1:n}, T_{p,\textrm{init}}, \textrm{DBMC}_{p,\textrm{init}}, \textrm{SF}), \\
& \sum_{d=1}^{n} (\textrm{MT}_d = 1) - 6 \leq 0 \quad \textrm{(Constraint 
 1)} \\
& 3 - \sum_{d=1}^{n} (\textrm{MT}_d = 2) \leq 0 \quad \textrm{(Constraint 
 2)} \\
& \sum_{d=2}^{n} (\textrm{MT}_d = 2 | \textrm{MT}_d = 3)|T_{a,d}-T_{a,d-1}| \leq 0 \quad \textrm{(Constraint 3)} \\
\text{With:} \quad & T_{a,d=1:n} \in \{0, 1, 2, 3, 4, 5, 6, 7, 8, 9, 10\}, \quad \textrm{MT}_{d=1:n} \in \{0, 1, 2, 3\}, \\
& v_a = \frac{\dot{V}_a}{\rho_a A_{\textrm{nozzle}}}, \quad \dot{V}_a = 500 \textrm{ SCFM}, \quad A_{\textrm{nozzle}} = \SI{0.0233}{\meter \squared}\\
& T_{p,\textrm{init}} = \SI{20}{\celsius}, \quad \textrm{DBMC}_{p,\textrm{init}} = 1.5
\end{aligned}
\end{equation}

Here, $f$ and $g$ are objective and constraint functions backed by the simulation environment. In practice, $f$ and $g$ take as input a vector of length $n$ that encodes $T_{a,d=1:n}$ (11 discrete levels) and $\textrm{MT}_{d=1:n}$ (4 discrete module types) where $n$ is the number of dryer modules used by the best solution found by RLCBS. The input can be constructed in a manner similar to the RL action space described in Table~\ref{table:states-actions}. Air velocity $v_a$ is determined by fan power (air volumetric flow rate), which is always kept at maximum, and $\rho_a$, which is in turn influenced by $T_{a,d}$. The machine speed, $v_{\textrm{m}}$, controls the total drying time and is manually specified for each optimization run by specifying the speed factor, $\textrm{SF}$. Conversion between $v_m$ and \textrm{SF} is as described in Equation~\ref{eqn:speed-factor-vs-machine-speed}, and all IR modules are disabled.

For NSGA-II, we use an implementation provided in \texttt{pymoo}~\citep{9078759}, a popular Python library for multi-objective optimization algorithms. We use population size of 32 and 100 generations for all experiments. We have also explored population size of 64 and 50 generations, but found the results to be much worse, as NSGA-II often fails to find any feasible solution. Keeping the population size a multiple of 32 allows the algorithm to take full advantage of 32 CPU threads provided by the AMD 5950X CPU in our experiment setup, thus minimizing the wall time required for solution. Although we use a Redis cache to serve all parallel worker threads and reduce duplicate simulation work, we found that the cache hit rate is low at 13.95\% (65,850 / 472,176) due to the random nature of the candidate mutation procedure in NSGA-II. Consequently, NSGA-II did not benefit significantly from the global cache. Each NSGA-II session takes approximately 0.5-\SI{1.5}{\hour} depending on the machine speed setting, with slower machine speed settings taking more time as the simulation timestep length in the physics-based model is fixed at \SI{0.25}{\ms} regardless of the total drying time.

\section{Results}\label{sec:results}
\begin{figure}
\centering
\includegraphics{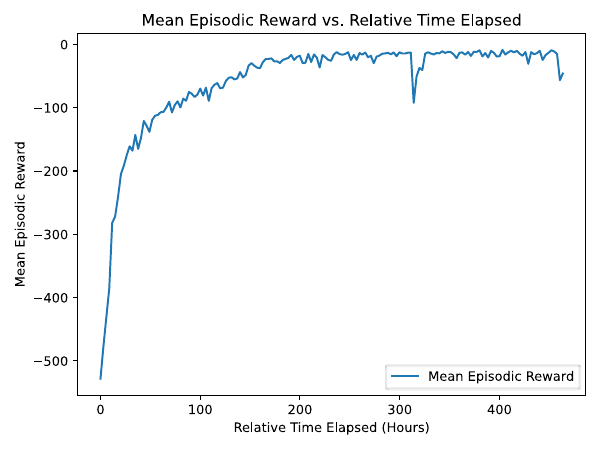}
\caption{Moving average of cumulative episodic reward over past 100 episodes for the PPO RL agent trained on the Smart Dryer simulation environment with randomized initial machine speed levels.}
\label{fig:lr-curve}
\end{figure}

The RL agent was trained using the PPO algorithm for 10 million timesteps in the Smart Dryer physics-based simulation environment. Training experiments were parallelized over 30 processes and took approximately 20 days of wall time on a PC with an AMD 5950X CPU and an Nvidia RTX3090 GPU. Physics-based simulation is performed on CPU and RL model training is performed on GPU. Due to fully randomized machine speed settings, the training process does not benefit from the Redis global cache. As shown in Figure~\ref{fig:lr-curve}, the RL agent started with a negative reward and made progress in the initial stages of training. However, after 5 million timesteps, the running mean of cumulative episodic reward plateaued below 0, indicating that the RL model could not outperform the SQP-optimized solution using greedy search.

Although machine speed levels are fully randomized during training, we use fixed machine speed levels between 0.25 and 0.75 in increments of 0.05 for evaluation, since the reward function in Equation~(\ref{eqn:reward-function}) requires comparing the drying outcome of the RL solution with the SQP baseline calculated at these grid points, while any position in between requires interpolation. With the test initial conditions aligned with precalculated positions, more accurate results can be reported as no interpolation error is introduced to the results. 

For each machine speed level, we run 8 sessions using RLCBS with increasing number of beams $n_b$: 2, 4, 8, 16, 32, 64, 128, 256. Due the presence of a Redis global cache, results from previous runs can be reused by subsequent runs, provided that the machine speed level and any other initial process conditions remain constant. Actual cache hit rates after finishing all CBS experiments up to $n_b = 256$ are 72.61\% (254,989 / 351,155) for Table~\ref{tab:results-3} and 65.06\% (194,112 / 298,336) for Table~\ref{tab:results-123}. Among the 8 runs for each machine speed level, we retrieve the lowest number of beams required to achieve the best energy savings across all sessions, and report the cumulative run time start from $n_b = 2$ to that beam size.

We report the performance of three methods: RL greedy search, RLCBS, and NSGA-II under minimal constraint (constraint 3 only) in Table~\ref{tab:results-3}. Empty results for RL greedy search indicates that generated sequence caused the simulation environment to fail, leading to a large negative reward ($R < -1000$). Reasons for failure include an excessive high temperature of air leading to boiling temperatures in the paper and excessive overdrying beyond the target DBMC. Across all machine speed levels, RLCBS achieves average energy savings of \SI{15.857}{\kjmm}, a \SI{28.962}{\kjmm} improvement over greedy search after excluding any greedy search failure cases. RLCBS slightly underperforms NSGA-II by \SI{0.979}{\kjmm}, although the time to solution is much faster, as it requires on average \SI{19.08}{\minute} for the 8 consecutive RLCBS runs using number of beams of 2 to 256, while NSGA-I requires on average \SI{49.17}{\minute} per optimization job. By adjusting the maximum number of beams used, RLCBS gives the users more flexibility in optimizing results within the allocated time budget. When the user consistently run optimization at max 256 beams, RLCBS can still provide significant speed improvement over NSGA-II.

\begin{table}
\centering
\caption{Comparison of greedy search, RLCBS, and NSGA-II performance under constraint 3. $R$ represents energy saved vs. a SQP-optimized baseline under constraints 1 and 3. Empty entries for greedy search indicate that the generated sequence had cause the simulation to abort and fail to return any result. Time reported for RLCBS is cumulative value from $n_b = 2$ to $n_b$ for best result.}
\label{tab:results-3}
\begin{tabular}{l|ll|llll|ll}
\hline
\textbf{} & \multicolumn{2}{l|}{\textbf{Greedy search}} & \multicolumn{4}{l|}{\textbf{RLCBS}} & \multicolumn{2}{l}{\textbf{NSGA-II}} \\
$v_{\textrm{m}}$ (\SI{}{\meter \per \second}) & $R$ (\SI{}{\kjmm}) & Time (s) & $n_b$ & $R$ (\SI{}{\kjmm}) & Cumul. time (s) & \# dryers & $R$ (\SI{}{\kjmm}) & Time (s) \\ \hline
0.0512 & -4.015 & 0.795 & 256 & 13.358 & 561.919 & 8 & -0.141 & 1611.535 \\
0.0482 & - & 0.86 & 64 & 14.876 & 305.387 & 11 & 25.915 & 2221.894 \\
0.0452 & -16.069 & 0.844 & 128 & 11.178 & 640.193 & 9 & 16.647 & 2127.023 \\
0.0423 & - & 0.867 & 256 & 11.658 & 1187.196 & 10 & 18.352 & 2526.841 \\
0.0393 & -0.568 & 0.868 & 256 & 10.048 & 1468.446 & 11 & 16.305 & 2726.594 \\
0.0363 & -16.735 & 0.86 & 32 & 14.51 & 360.823 & 10 & 13.415 & 2871.836 \\
0.0334 & - & 0.943 & 256 & 13.058 & 1981.65 & 10 & 17.927 & 3020.567 \\
0.0304 & -18.513 & 0.866 & 4 & 18.0 & 111.892 & 10 & 17.286 & 3200.691 \\
0.0274 & - & 0.923 & 128 & 18.493 & 1305.378 & 12 & 16.268 & 3882.45 \\
0.0244 & -22.727 & 0.866 & 256 & 22.681 & 3143.369 & 10 & 18.987 & 3594.991 \\
0.0215 & - & 0.869 & 64 & 26.572 & 1525.45 & 12 & 24.230 & 4666.397 \\
Average & -13.105 & 0.869 & - & 15.857 & 1144.70 & - & 16.836 & 2950.074 \\ \hline
\end{tabular}
\end{table}

With all constraints 1, 2, 3 applied, it is no longer possible to use RL greedy search, as satisfaction of constraint 2 cannot be guaranteed in greedy search. A comparison of RLCBS and NSGA-II performance is provided in Table~\ref{tab:results-123}. As a result of additional constraints, both RLCBS and NSGA-II saw reduction in energy saved vs. the baseline. However, it should be noted that the dryer module sequence configuration used by the SQP baseline violates constraint 2, and the relative energy savings in this case should only be used as a metric for relative performance between RLCBS and NSGA-II.
We do observe that RLCBS adapted to additional constraints better than NSGA-II as it suffered less from performance decrease. Across all machine speed levels, RLCBS with final timestep refinement achieves average energy savings of \SI{6.283}{\kjmm}, outperforming NSGA-II by \SI{0.062}{\kjmm}. Average time to solution is \SI{8.16}{\minute}, a 6.64-fold improvement over NSGA-II.

\begin{table}
\centering
\caption{Comparison of RLCBS, and NSGA-II performance under constraints 1, 2, and 3. $R$ represents energy saved vs. a SQP-optimized baseline under constraints 1 and 3. Time reported for RLCBS is cumulative value from $n_b = 2$ to $n_b$ for best result.}
\label{tab:results-123}
\begin{tabular}{l|llll|ll}
\hline
\textbf{} & \multicolumn{4}{l|}{\textbf{RLCBS}} & \multicolumn{2}{l}{\textbf{NSGA-II}} \\
$v_{\textrm{m}}$ (\SI{}{\meter \per \second}) & $n_b$ & $R$ (\SI{}{\kjmm}) & Cumul. time (s) & \# dryers & $R$ (\SI{}{\kjmm}) & Time (s) \\ \hline
0.0512 & 64 & 14.69 & 349.081 & 11 & 11.126 & 2234.97 \\
0.0482 & 2 & 10.505 & 44.009 & 11 & 9.355 & 2160.38 \\
0.0452 & 64 & 8.113 & 472.658 & 12 & -1.583 & 2688.946 \\
0.0423 & 128 & 4.87 & 746.583 & 11 & 0.957 & 2469.21 \\
0.0393 & 4 & 3.351 & 121.811 & 12 & 10.116 & 3348.631 \\
0.0363 & 4 & -4.165 & 134.12 & 11 & -5.756 & 2840.519 \\
0.0334 & 8 & -1.912 & 201.514 & 12 & 5.584 & 3693.685 \\
0.0304 & 2 & -3.399 & 73.087 & 12 & 5.892 & 3716.842 \\
0.0274 & 256 & 8.703 & 2323.453 & 12 & 1.399 & 4178.648 \\
0.0244 & 2 & 10.073 & 92.18 & 12 & 14.557 & 4625.78 \\
0.0215 & 32 & 18.28 & 824.812 & 11 & 16.783 & 3783.635 \\
Average & - & 6.283 & 489.392 & - & 6.221 & 3249.204 \\ \hline
\end{tabular}
\end{table}

\section{Discussion}
\label{sec:discussion}
\subsection{Sample Trajectory}
\begin{figure}
\centering
\includegraphics[width=0.8\columnwidth]{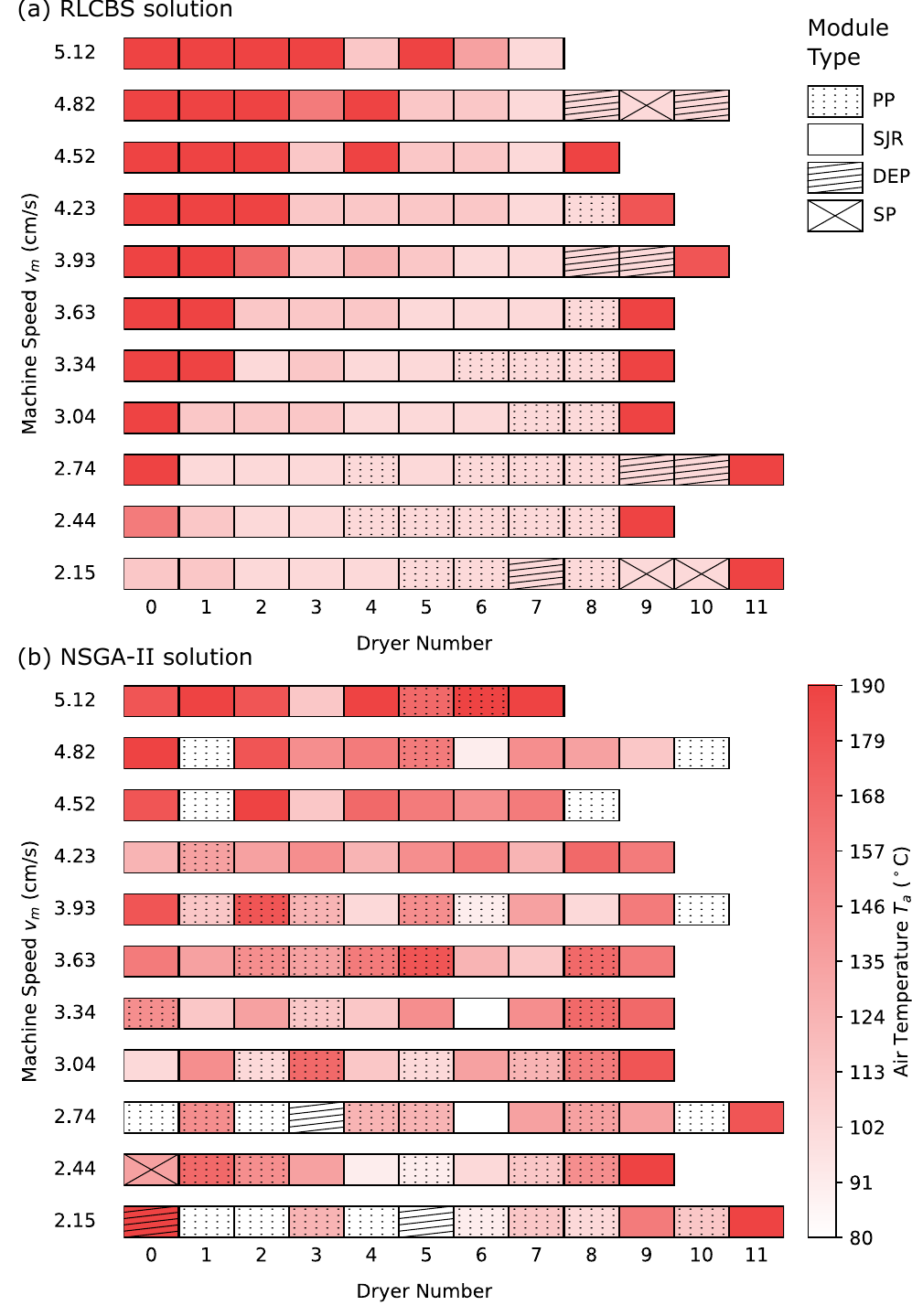}
\caption{Comparison of RLCBS and NSGA-II solutions under constraint 3.}
\label{fig:visualization_3}
\end{figure}

\begin{figure}
\centering
\includegraphics[width=0.8\columnwidth]{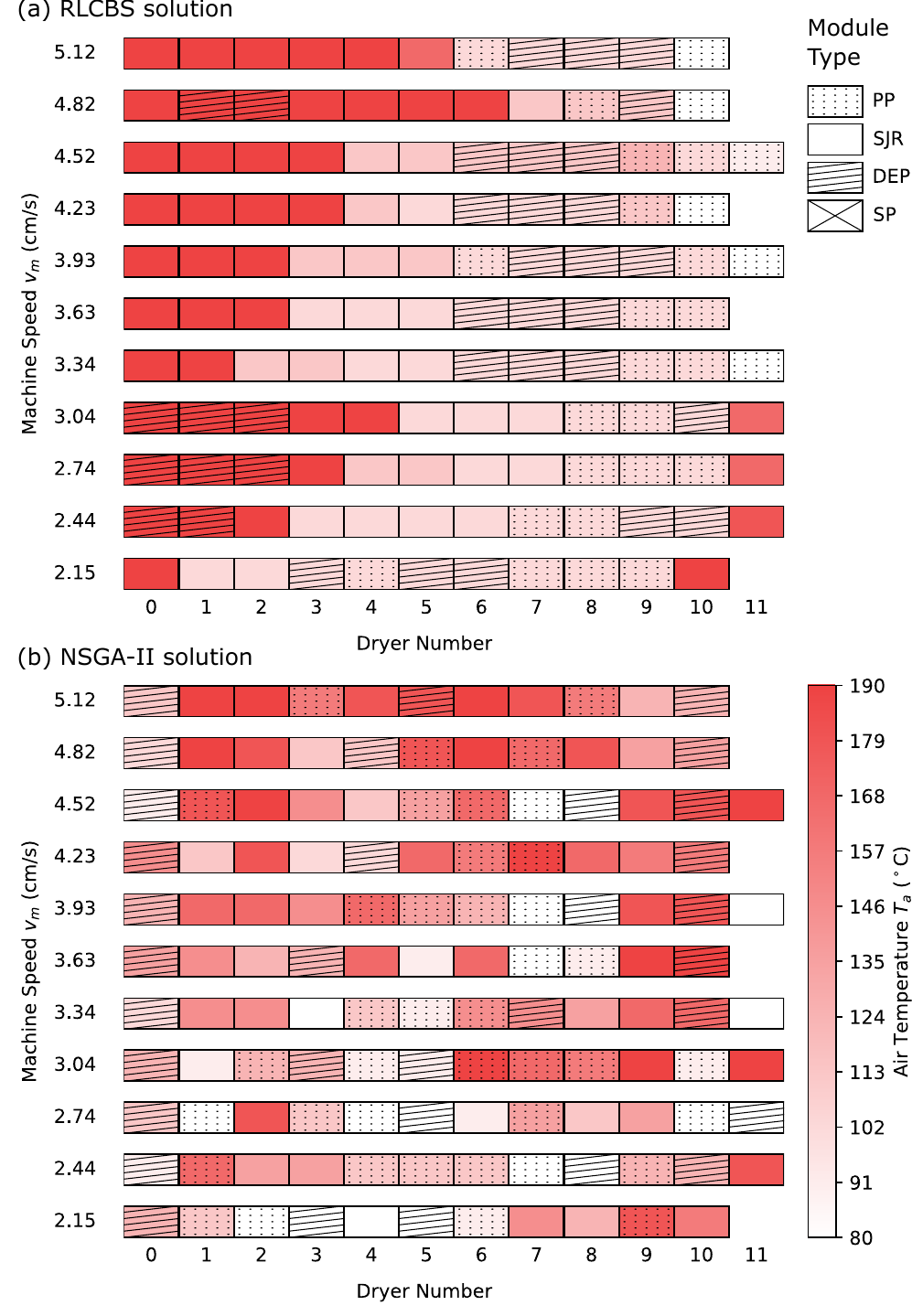}
\caption{Comparison of RLCBS and NSGA-II solutions under constraints 1, 2, and 3.}
\label{fig:visualization_123}
\end{figure}

To study how constraints are implemented by RLCBS and NSGA-II in their solution sequences, we visualize the solutions generated by both methods for the experiments described in Section~\ref{sec:results}. Figure~\ref{fig:visualization_3} corresponds to Table~\ref{tab:results-3}, while Figure~\ref{fig:visualization_123} corresponds to Table~\ref{tab:results-123}. In these figures, each row represents a solution sequence containing up to 12 dryer modules for a specific process operating condition. The four types of dryer modules are differentiated by patterns, and the fill color indicates the air supply temperature at each location.

RLCBS demonstrates a clear strategy in allocating dryer modules and assigning temperature profiles across different operating conditions, likely due to prior knowledge learned by the RL policy in the unconstrained environment. In contrast, NSGA-II relies on random mutation for population evolution, and it is difficult to generalize high-level strategies from the solutions. For both sets of experiments, RLCBS consistently applies higher $T_a$ in the initial dryer module locations, gradually reducing $T_a$ as the paper approaches the DBMC target. As machine speed $v_m$ decreases, the algorithm reduces the number of high-temperature modules at the beginning of the sequence to account for increased total drying time and to avoid overdrying. Towards the end of drying, RLCBS applies high temperature to the final module position to achieve the desired DBMC target.

In terms of module type selection, RLCBS demonstrates a clear preference for SJR over PP modules, particularly under low-residence time settings where drying capacity requirements are high. This preference is especially evident in Figure~\ref{fig:visualization_3}, where no constraints are placed on the maximum number of SJR modules in a drying sequence. Here, RLCBS uses almost exclusively SJR units, only switching to PP or DEP units when total drying times increase or when the paper approaches the end of the drying process. This preference is justified by the superior drying capacities of SJR compared to other module types.

In Figure~\ref{fig:visualization_123}, where the total number of SJR modules is limited to 6, RLCBS adapts by placing SJR units primarily at the beginning of the drying process and utilizing PP units, which have the second-highest drying capacity, towards the end of the process.

Despite the DEP modules' low energy consumption, they have relatively low drying capacity due to the absent of forced air flow and the reliance on natural convection for drying at DEP and SP module locations. The RLCBS algorithm predominantly employs DEP modules towards the end of the drying process, although in several instances with extended residence times, RLCBS also applies DEP modules at the start of drying, coupled with high air temperatures.

The DEP module placement behaviors differ significantly between RLCBS and NSGA-II. In Figure~\ref{fig:visualization_3}, where no constraints mandate DEP usage, NSGA-II rarely incorporates these modules to the solution sequence. Conversely, in Figure~\ref{fig:visualization_123}, which requires the inclusion of at least 3 DEP units, NSGA-II's placement of DEP modules appears largely random, except that the first dryer is always a DEP unit.

The consistent placement of a DEP module in the first dryer position by NSGA-II is likely not motivated by energy savings. Instead, it may be attributed to the exception in constraints applied to the first module position. Specifically, constraint 3, which enforces equal $T_a$ for DEP and SP modules relative to the preceding module, does not apply to the first position. Consequently, consistently placing a DEP module in this position helps NSGA-II to reduce constraint violations in its solutions.

\subsection{Comparison of SGBS, RLGBS, RLCBS}
Comparison between RL-guided beam search methods supporting only negative constraints, SGBS~\citep{neurips2022_39b9b60f} and RLGBS, shows different approaches in beam candidate selections. Namely, SGBS is simulation-guided in that, although RL model is responsible for growing the search tree, pruning of beam candidates relies on simulation environment feedback. Conversely, RLGBS relied on RL policy to grow and prune beam candidates. 

SGBS may be more robust in that beam candidates selected at each timestep are guaranteed to be the best-performing beams in terms of cumulative reward. In the case of an RL model operating on a problem with a different distribution, SGBS is more likely to perform well due to consistant feedback from the environment \emph{so far}. However, due to its reliance on reward feedback for beam pruning, SGBS is limited to tasks with dense reward. Additionally, even for dense reward tasks, it cannot be guaranteed that selecting beam candidates based on cumulative reward at intermediate timesteps will lead to a globally optimal solution.

RLGBS does not rely on reward feedback at all during search process, only that the best returned process at the end of search is selected based on cumulative reward. This strategy allowed it to operate on sparse reward tasks. However, as mentioned by the SGBS authors, relying solely on the RL policy for exploration may not be the most efficient approach. This is partially remedied by the large beam sizes used by RLGBS (up to 256). 

In developing RLCBS, we attempt to strike a compromise between the two approaches. RLCBS still relies on the RL policy for beam growth and partly for beam candidate selection, but there are additional pruning at beam selection time to ensure that any failed beams (beam candidates that result in the simulation environment to arrive at a truncated state with negative reward) are not passed to the next timestep. When finalizing beams, similar to RLGBS, we evaluate all completed hypotheses and return the best hypothesis based on cumulative reward. Finally, we attempt to refine a small number of returned hypotheses (1 for $n_b \leq 4$, 4 for $n_b > 4$) by swapping the last action of the sequence with every possible action, discard any sequence that results in violation of constraints, and re-evaluate the remaining sequences using the simulation environment. Doing so requires evaluating at most $4 \times |\mathcal{A}| = 176$ new sequences for any beam size, and therefore does not affect the overall runtime complexity of the algorithm.

\subsection{Energy Savings}
Compared to average energy consumption baselines set by SQP in Table~\ref{tab:sqp-baseline-values}, average energy savings by RLCBS are 1.866\% under constraint 3, and 0.737\% under constraint 1, 2, 3. Although percentage savings are low, this is partly due to the baseline is already optimized by SQP. It proved difficult to provide a single-stage baseline using any na\"ive policy, as due to the variation in the total drying time (2-\SI{5}{\minute}), any single-stage, constant parameter policy will fail in most test cases due to underdrying or severe overdrying. Additionally, SQP being able to operate in the continuous domain for the optimized variables may have been advantageous vs. RLCBS and NSGA-II under their current configuration of 11 discrete air temperature levels. A potential topic of future research may be to adapt RLCBS to RL agents with continuous action spaces.

Despite the low percentage savings, the potential contribution to sustainability is substantial when considering the scale of application of paper manufacturing. Taking into account the rate of thermal energy consumption of \SI{21}{\tera\joule} per day~\citep{osti_1335587}, the percentage of energy savings could result in a reduction of more than \SI{0.1}{\tera\joule} in thermal energy daily. However, it should be noted that energy savings can vary due to variations in operating conditions, process configuration, and paper grade.

\subsection{Limitations}
By exploring multiple beam candidates in parallel and considering the overall sequence feasibility over all timesteps, real-time performance is sacrificed by replacing greedy search with beam search. However, this trade-off may be justified in problems where the cost of mistakes is high and real-time performance is not a critical requirement. For the paper drying process considered in this work, frequent adjustments to the control parameters are necessary to adapt to variations in operating conditions. Yet, real-time performance is not essential as these parameter adjustments may only be required on a batch or campaign basis, or the system's response time may not be fast enough to warrant real-time control. In such cases, employing RLGBS to obtain higher cumulative episodic rewards may be desirable despite the increased computational cost compared to greedy decoding. Conversely, in time-critical problems or problems involving non-deterministic environments, alternative methods like greedy search or one-step lookahead may be more appropriate, or the limitation could be circumvented by training a new RL agent using imitation learning on a dataset of high-quality sequences generated by RLGBS.

Additionally, the beam search decoder proposed in this work requires a copy of the simulation environment to determine the next states for sequence continuation and checking stopping conditions. In practice, such an oracle simulation environment may not always be available. A possible solution is to train a surrogate model of the environment and use the surrogate in place of the oracle during decoding. Further investigations on the impact of discrepancies between the first-principle oracle and surrogate models on the performance of generated actions may be needed.

Finally, the simulation environment introduced in this work makes simplifying assumptions to the problem setup, including discrete temperature levels and assuming uniform air distribution to all drying modules. The energy savings are realized by adjusting the temperature of the hot air supply in a way that may not be feasible in the physical setup, and this study is not concerned with the potential impact of optimally selected dryer module and air temperature sequence configurations on the properties of the paper. A process controller implemented in a real paper machine should also ensure that product quality constraints are met. A potential direction for further research may be to impose quality targets as constraints while reducing the energy consumption of the process.

\section{Conclusion}
\label{sec:conclusion}
We present Reinforcement Learning Constrained Beam Search (RLCBS) as a method to incorporate complex design constraints to RL-generated actions at inference time. RLCBS enables policy-based RL methods to search for high-quality solutions efficiently in exponential search spaces, while injecting design constraints unseen at training time. Our experiments on optimization of process configuration of a novel, modular Smart Dryer test bed show that RLCBS adapts to complex inference-time design constraints, yielding similar or better performance than NSGA-II under the same set of constraints, while providing a 2.58-fold or higher speed advantage and potentially allowing flexibility in trading off between solution time and performance. RLCBS could benefit various RL-based optimization problems where real-time performance is not a critical requirement.

\section*{Acknowledgement}
This study was financially supported by the {U.S.} Department of Energy, Office of Advanced Manufacturing under Award Number DE-EE0009125, the Massachusetts Clean Energy Center (MassCEC), and Center for Advanced Research in Drying (CARD), a {U.S.} National Science Foundation Industry-University Cooperative Research Center. CARD is located at Worcester Polytechnic Institute and the University of Illinois at Urbana-Champaign (co-site). The views expressed herein do not necessarily represent the views of the U.S. Department of Energy or the United States Government.

\bibliographystyle{unsrtnat}
\bibliography{template}  

\end{document}